\documentclass[]{gensi}
\usepackage{tabularx}
\usepackage[toc,page,header]{appendix}

\DeclareUnicodeCharacter{FF1A}{:}
\DeclareUnicodeCharacter{2032}{:}
\PassOptionsToPackage{float}{none}
\usepackage{floatrow}
\newfloatcommand{capbtabbox}{table}[][\FBwidth]

\usepackage{minitoc}
\usepackage{cleveref} 
\usepackage{microtype}
\usepackage{subcaption}
\usepackage{amsmath} 
\usepackage{amssymb} 
\usepackage{amsfonts}       
\usepackage{pgfplots}
\usepackage{pgfplotstable}
\usepackage{xcolor}
\usepackage{CJKutf8}
\usetikzlibrary{patterns}
\usepackage{multirow}
\usepackage{setspace}
\usepackage{xcolor}
\usepackage{hyperref}
\tcbuselibrary{theorems}
\usepackage{float}
\usepackage{caption}
\usepackage{listings}
\usepackage{wrapfig}
\usepackage{xspace}
\usepackage{graphicx}      
\usepackage{subcaption}    
\usepackage{tcolorbox}

\usepackage{fix-cm}

\usepackage{algorithm}
\usepackage{algorithmic}
\usepackage{mathtools}
\usepackage{amsthm}

\theoremstyle{plain}
\newtheorem{theorem}{Theorem}[section]
\newtheorem{proposition}[theorem]{Proposition}

\theoremstyle{definition}
\newtheorem{definition}[theorem]{Definition}

\theoremstyle{remark}

\usepackage{natbib}
\usepackage{latexsym}

\usepackage{url}
\usepackage{amssymb}
\usepackage[utf8]{inputenc}
\usepackage{microtype}
\usepackage{booktabs}
\usepackage{pifont} 
\usepackage{multirow}
\usepackage{makecell}
\usepackage{paralist}
\usepackage{xspace}
\usepackage{color}
\usepackage{xcolor}
\usepackage{colortbl}
\usepackage{adjustbox}
\usepackage{hyperref} 
\usepackage[edges]{forest}
\usepackage{tikz} 
\usepackage{caption}
\usepackage{amsfonts}

\hypersetup{
    colorlinks,
    linkcolor={blue!80!black},
    citecolor={blue!80!black},
}
\tikzset{
    root/.style =             {align=center, text width=1cm, rounded corners=3pt, line width=0.3mm, fill=gray!10, draw=gray!80, font=\small},
    demographic/.style =         {align=center, text width=1.8cm, rounded corners=3pt, line width=0.3mm, fill=blue!10, draw=blue!80, font=\footnotesize},
    demographic_work/.style =    {align=center, text width=10cm, rounded corners=3pt, line width=0.3mm, fill=blue!10, draw=blue!0, font=\footnotesize},
    character/.style =         {align=center, text width=1.8cm, rounded corners=3pt, line width=0.3mm, fill=red!10, draw=red!80, font=\footnotesize},
    character_work/.style =    {align=center, text width=10cm, rounded corners=3pt, line width=0.3mm, fill=red!10, draw=red!0, font=\footnotesize},
    personalization/.style =           {align=center, text width=1.8cm, rounded corners=3pt, line width=0.3mm, fill=cyan!10, draw=cyan!80, font=\footnotesize},
    personalization_work/.style =      {align=center, text width=10cm, rounded corners=3pt, line width=0.3mm, fill=cyan!10, draw=cyan!0, font=\footnotesize},
    risk/.style =         {align=center, text width=1.8cm, rounded corners=3pt, line width=0.3mm, fill=orange!10, draw=orange!80, font=\footnotesize},
    risk_work/.style =    {align=center, text width=10cm, rounded corners=3pt, line width=0.3mm, fill=orange!10, draw=orange!0, font=\footnotesize},
}

\usepackage{CJK}

\def\method{{SLM}}
\def\methodsy{{SLM }}

\definecolor{mygray}{gray}{.9}

\title{
    ShortListing Model: A Streamlined Simplex Diffusion for Discrete Variable Generation
}

\author[1,2,3,*]{Yuxuan Song}
\author[1,3,*]{Zhe Zhang}
\author[1,3,*]{Yu Pei}
\author[1,3]{Jingjing Gong}   
\author[1,2,3]{Qiying Yu}
\author[2]{Zheng Zhang}
\author[2]{Mingxuan Wang}
\author[1,3,\dagger]{Hao Zhou}
\author[1,3]{Jingjing Liu}
\author[1,3]{Wei-Ying Ma}

\affiliation[1]{Generative Symbolic Intelligence Lab (GenSI), Tsinghua University }
\affiliation[2]{ByteDance Seed}
\affiliation[3]{Institute for AI Industry Research (AIR), Tsinghua University}

\contribution[*]{Equal Contribution}
\contribution[\dagger]{Corresponding authors: zhouhao@air.tsinghua.edu.cn}

\abstract{
    Generative modeling of discrete variables is challenging yet crucial for applications in natural language processing and biological sequence design. We introduce the Shortlisting Model (\method), a novel simplex-based diffusion model inspired by progressive candidate pruning. SLM operates on simplex centroids, reducing generation complexity and enhancing scalability. Additionally, SLM incorporates a flexible implementation of classifier-free guidance, enhancing unconditional generation performance. 
Extensive experiments on DNA promoter and enhancer design, protein design, character-level and large-vocabulary language modeling demonstrate the competitive performance and strong potential of SLM. Our code can be found at \url{https://github.com/GenSI-THUAIR/SLM}
}

\begin{document}

    \maketitle

    \vspace{-16mm}


\section{Introduction}
\label{sec:intro}

\begin{wrapfigure}{r}{0.4\textwidth}
    \vskip 5pt
  \begin{center}
    \includegraphics[width=0.98\textwidth]{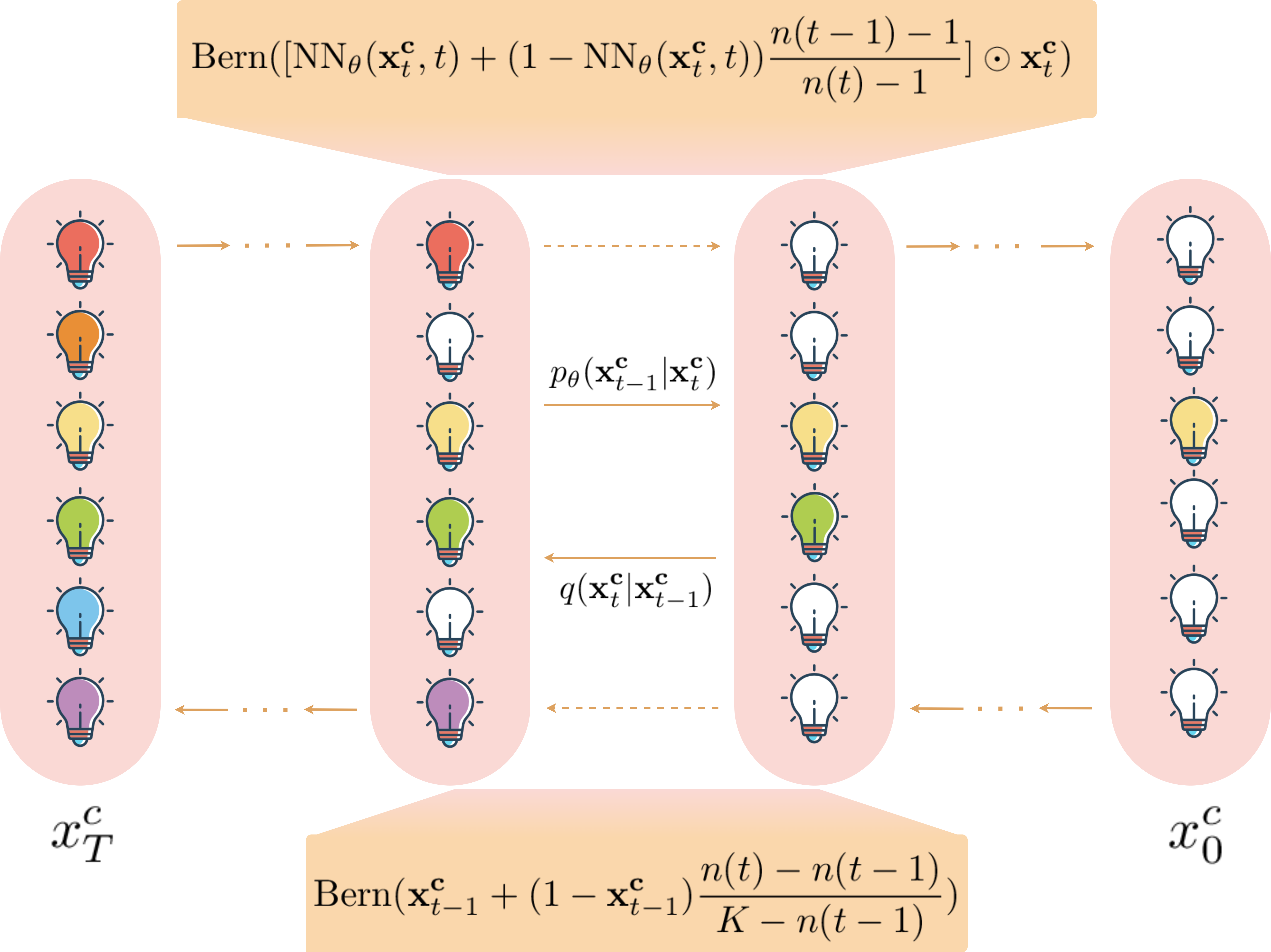}
  \end{center}
  \caption{\method's forward and reverse process. Comparisons between MDLM and DP3M-Uniform is located in Appendix \ref{app:cmp_visual}}
  \label{fig:overview}
  \vskip 10pt
\end{wrapfigure}

Autoregressive models such as large language models (LLMs), although having achieved remarkable success in text generation~\citep{achiam2023gpt,brown2020language}, struggle over tasks that lack an intrinsic sequential-ordering inductive bias, such as DNA design~\citep{avdeyev2023dirichlet,stark2024dirichlet}, protein sequence design~\citep{wang2024diffusion,lin2023evolutionary} and molecular graph generation~\citep{vignac2022digress}. Consequently, there is growing interest in developing new paradigms for discrete variable generation, such as diffusion-based~\citep{loudiscrete,sahoo2024simple,shi2024simplified} and flow-matching-based~\citep{gat2024discrete,davis2024fisher,cheng2024categorical} methods.

Recent discrete generative models are generally classified into two main categories based on their operational spaces: discrete-space models and continuous-space models. The former, specifically discrete diffusion model~\citep{loudiscrete,xu2024energy}, mimics continuous diffusion processes using substitution or masking operations to decompose information, which has shown impressive performance for discrete generative modeling.
However, these discrete counterparts differ fundamentally from original continuous diffusion models~\citep{ho2020denoising}, where the smooth and gradual information transitions intrinsic to the generation process are key to their success. On the other hand, continuous-space models map discrete data into continuous representations.
While benefiting from continuous properties, capturing the geometric structure and adhering to constraints via continuous generative models introduces new challenges. In this context, simplex-based approaches~\citep{cheng2024categorical,davis2024fisher,graves2023bayesian} offer a balanced solution by representing discrete data on the probability simplex, which naturally adheres to the fundamental properties of categorical distributions.

However, existing simplex-based approaches often rely on intricate operations to define trajectories over the entire continuous space.
For instance, Statistical Flow Matching(SFM)~\citep{cheng2024categorical} and Fisher Flow Matching~\citep{davis2024fisher} define geodesics based on sphere map and the Fisher-Rao metric. The training also incorporates Riemannian optimal transport. Similarly, Bayesian Flow Networks (BFNs)~\citep{graves2023bayesian} involve heavy mathematical derivations and change-of-variable techniques to define simplex trajectories through Gaussian-formed count variables. 
Despite mathematical rigor, their demanding complexity limits scalability in large-scale generative tasks.

\begin{figure*}[t]
\begin{center}
\includegraphics[width=0.9\textwidth]{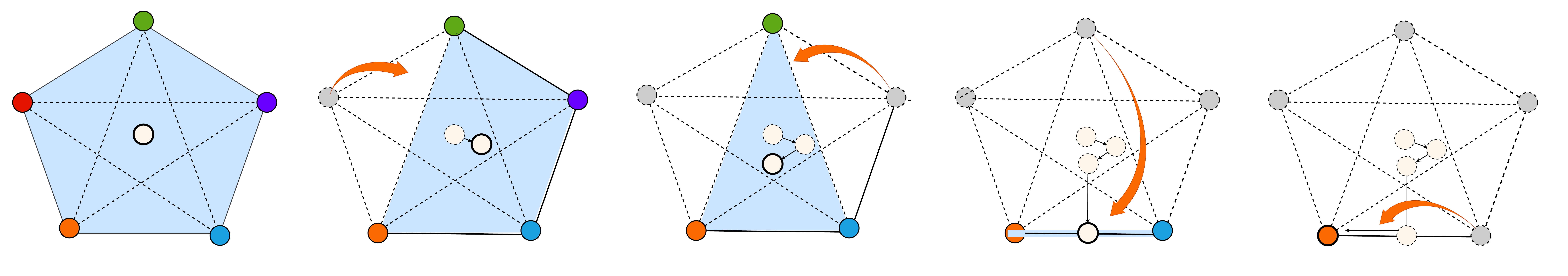}
\end{center}
\caption{Pathological behavior of \methodsy on one simplex with K = 5($\Delta^4$). Each vertex represents one of the categorical targets while the trajectory of the white point serves as a probability path in sampling. Note that the trajectory of shortlisting model could be seen as jumping among the centroids of subspaces in simplex space. 
}
\label{fig:SLM_simplex}
\end{figure*}

In this paper, we aim to preserve the core principle of simplex-based methods, \textit{gradual information growth}, while exploring simpler yet effective alternatives. We propose viewing discrete variable generation as progressive candidate pruning, starting from the full category set and iteratively narrowing down to a single choice. We term this approach \textbf{ShortListing Models}(SLM). Formally, shortlisting models reside within the diffusion framework, trainable via the variational lower bound (VLB). In contrast to existing simplex-based methods using vocabulary-level MSE loss~\citep{graves2023bayesian,cheng2024categorical,davis2024fisher}, our approach employs a simplified cross-entropy objective, effectively mitigating vanishing gradient issues and better handling large-vocabulary settings. As illustrated in Figures~\ref{fig:overview} and~\ref{fig:SLM_simplex}, our model effectively reduces degrees of freedom by modeling transitions among the simplex centroids rather than the entire simplex. Additionally, shortlisting models offer flexibility for adaptations such as classifier-free guidance.

We comprehensively evaluate the proposed approach over various discrete generation tasks and benchmarks, including char-level language modeling, large-scale language modeling, DNA promoter and enhancer design, and Protein sequence design. Specifically, we achieve strong performance among non-autoregressive methods on text8 and obtain competitive results on OpenWebText, where previous simplex-based approaches had difficulty generating reasonable outputs. In DNA design tasks, our non-guided variants achieve state-of-the-art (SOTA) results. Furthermore, with classifier-free guidance, our model attains stronger performance while remaining less sensitive to hyperparameters.
Additionally, our 38M-parameter shortlisting model can design proteins with enhanced foldability, fitness, self-consistency and diversity, surpassing the larger, well-known ESM2-150M model~\citep{lin2022language}.

    \section{Preliminary}

\subsection{Definition and Notations}

We encode a discrete variable with $K$ distinct categories using one-hot vectors $\mathbf{e}=\left[e_1, e_2, \ldots, e_K\right]^T$. In each vector, only the $i$-th element $\mathbf{e}(i)=1$ signifies the inclusion of the $i$-th category, while all other elements are zero.
\begin{definition}
\label{def:cs}
    A \textbf{candidate set} for $K$ categories is defined as a binary-valued vector $\mathbf{c}=\left[c_1, c_2, \ldots, c_K\right]^T$, where each $c_i \in\{0,1\}$, and the vector has at least one non-zero entry, \emph{i.e.,} $\mathbf{1}^T \mathbf{c} > 0$.
\end{definition}
\vspace{-5pt}
The candidate-set variable $\mathbf{c}$ encodes the selection status of each category: $\mathbf{c}(i)=1$ indicates that the $i$-th category is included, while $\mathbf{c}(i)=0$ denotes its exclusion. Specifically, there are two distinct instances of the candidate variable: $\mathbf{c}$ is an all-one vector ($[1,\cdots,1]$) that represents maximum candidates setting, \emph{i.e.} all $K$ categories are selected; one-hot vector is another special case which represents minimum candidates setting, with only one category included.

\subsection{Diffusion Models}
Shortlisting model is a variant of diffusion models, 
which can be viewed as latent variable models where the latent variables form a Markov chain. Specifically, for a diffusion model with the sequence latent variable $\boldsymbol{x}_{1:T} = \boldsymbol{x}_1, \cdots \boldsymbol{x}_T$, the implied density function $p_\theta$ holds the following Markovness by definition:
$p_\theta(\boldsymbol{x}_0,\boldsymbol{x}_{1:T}) =  p_\theta(\boldsymbol{x}_0| \boldsymbol{x}_1) p_\theta(\boldsymbol{x}_1|\boldsymbol{x}_2) \cdots p_\theta(\boldsymbol{x}_{T-1}|\boldsymbol{x}_T)$.
To learn this latent variable model, a carefully designed constant variational distribution $q\left(\boldsymbol{x}_{1: T} \mid \boldsymbol{x}_0\right)=\prod_{t=1}^T q\left(\boldsymbol{x}_t \mid \boldsymbol{x}_{t-1}\right)$, referred to as the `forward process', is involved. Based on the variational distribution, the diffusion model is generally trained with the following variational lower bound~\citep{austin2021structured,ho2020denoising}:
\begin{align}
\label{eq:vlb}
L_{\mathrm{vlb}}=&\mathbb{E}_{q\left(\boldsymbol{x}_0\right)}[ \underbrace{D_{\mathrm{KL}}\left[q\left(\boldsymbol{x}_T \mid \boldsymbol{x}_0\right) \| p\left(\boldsymbol{x}_T\right)\right]}_{L_T}+ \sum_{t=2}^T \underbrace{\mathbb{E}_{q\left(\boldsymbol{x}_t \mid \boldsymbol{x}_0\right)}\left[D_{\mathrm{KL}}\left[q\left(\boldsymbol{x}_{t-1} \mid \boldsymbol{x}_t, \boldsymbol{x}_0\right) \| p_\theta\left(\boldsymbol{x}_{t-1} \mid \boldsymbol{x}_t\right)\right]\right]}_{L_{t-1}} \nonumber \\ 
& \underbrace{-\mathbb{E}_{q\left(\boldsymbol{x}_1 \mid \boldsymbol{x}_0\right)}\left[\log p_\theta\left(\boldsymbol{x}_0 \mid \boldsymbol{x}_1\right)\right]}_{L_0}] .
\end{align}
Here $q(\boldsymbol{x}_0)$ refers to the data distribution. 

    \section{Shortlisting Models}

Inspired by progressive candidate pruning, we effectively translate the generation of discrete variables into a category selection process, which begins by considering all categories in the vocabulary as potential candidates and progressively narrows down the options until reaching a final one-hot representation, indicating a single category. This section introduces the detailed components of the proposed shortlisting model as well as the training and sampling processes.

\subsection{Forward Candidate Appending}
We design a forward candidate appending process over the space of \textbf{candidate set} as introduced in Definition.~\ref{def:cs}. For any discrete variable $\mathbf{x}$, its one-hot representation (as a special case of the candidate set) is considered as the initial step, denoted as $\mathbf{x}^{\mathbf{c}}_0$. For the last step, $\mathbf{x}^{\mathbf{c}}_T$, we make it into an all-one vector ($\mathbf{1}$). Then we seek the following Markov chain interpolation $\mathbf{x}^{\mathbf{c}}_0 - \mathbf{x}^{\mathbf{c}}_1 -\cdots -\mathbf{x}^{\mathbf{c}}_T $ between  $\mathbf{x}^{\mathbf{c}}_0$ and $\mathbf{x}^{\mathbf{c}}_T$, which satisfies:
\begin{align}
\label{eq:markov}
\forall_{0 \leq i < j \leq T}~~\mathbf{1}^T\mathbf{x}^{\mathbf{c}}_i\leq \mathbf{1}^T\mathbf{x}^{\mathbf{c}}_j, [{\mathbf{x}^{\mathbf{c}}_j}]^T\mathbf{x}^{\mathbf{c}}_i = \mathbf{1}^T\mathbf{x}^{\mathbf{c}}_i
\end{align}
Recall $\mathbf{x}^{\mathbf{c}}$ is a binary-valued vector, hence the above condition essentially indicates the possible categories implied by the candidate set of early steps are \textbf{strictly scooped} by later steps.
We propose using a multivariate Bernoulli distribution to model the forward process over the candidate-set variable, denoted as $\mathbf{x}^{\mathbf{c}} \sim \operatorname{Bern}(\boldsymbol{\phi})$, where $\boldsymbol{\phi}$ is a $K$ dimensional vector representing the parameters of the distribution. 
To control the noise level, we introduce $n(t)$ as a scheduling function over the candidate numbers, where $1 \leq n(t) \leq K$. 
By our definition, $n(t)$ is a monotonically increasing function from time step 0 to $T$, designed to gradually perturb the signal. Intuitively, $n(t)$ can be viewed as controlling the number of ones in the vector $\mathbf{x}_t^{\mathbf{c}}$, representing the number of possible categories at time $t$. To satisfy the condition in Eq.~\ref{eq:markov}, we set $n(0)=1$ and $n(T)=K$, and define the transition probabilities from $t-1$ to $t$ as:
\begin{align}
\label{eq:trans}
    q(\mathbf{x}^{\mathbf{c}}_t|\mathbf{x}^{\mathbf{c}}_{t-1}) = \operatorname{Bern}\left(\mathbf{x}^{\mathbf{c}}_{t-1} + (1-\mathbf{x}^{\mathbf{c}}_{t-1})\frac{n(t)-n(t-1)}{K-n(t-1)}\right)
\end{align}
\begin{proposition}
    With Eq.~\ref{eq:trans} as the transition probability, the marginal distribution is defined as:
\begin{align}
\label{eq:marginal}
    q(\mathbf{x}^{\mathbf{c}}_t|\mathbf{x}^{\mathbf{c}}_{0}) = \operatorname{Bern}\left(\mathbf{x}^{\mathbf{c}}_{t-1} + (1-\mathbf{x}^{\mathbf{c}}_{t-1})\frac{n(t)-1}{K-1}\right)
\end{align}
 and corresponding posterior distribution $q(\mathbf{x}^{\mathbf{c}}_{t-1}|\mathbf{x}^{\mathbf{c}}_{t},\mathbf{x}^{\mathbf{c}}_{0}) $ also lies in the form of Bernoulli distribution, the analytical form of which is ($t \geq 2$):
 \begin{align}
 \label{eq:posterior}
    q(\mathbf{x}^{\mathbf{c}}_{t-1}|\mathbf{x}^{\mathbf{c}}_{t},\mathbf{x}^{\mathbf{c}}_{0})= \operatorname{Bern}\left(\mathbf{x}^{\mathbf{c}}_{0} + [(1-\mathbf{x}^{\mathbf{c}}_{0})\odot\mathbf{x}^{\mathbf{c}}_{t}] \frac{n(t-1)-1}{n(t)-1}\right)
 \end{align}
 Here $\odot$ stands for the Hadamard products between two vectors. 
\end{proposition}
Detailed proof can be found in Appendix \ref{app:proof}.

\subsection{Reverse Candidate Pruning}

\label{sec:reverse}
The reverse process implied by $p_\theta(\mathbf{x}^{\mathbf{c}}_{t-1}|\mathbf{x}^{\mathbf{c}}_{t})$ corresponds to the progressive candidate pruning process. We follow previous literature~\citep{austin2021structured,sahoo2024simple} to parameterize $p_\theta(\mathbf{x}^{\mathbf{c}}_{t-1}|\mathbf{x}^{\mathbf{c}}_{t})$, by combining a neural network($\theta$) predicted $\mathbf{x}^{\mathbf{c}}_{0}$ based on $\mathbf{x}^{\mathbf{c}}_{t}$ and the formulation of the posterior in Eq.~\ref{eq:posterior}:
\begin{align}
\label{eq:predicted}
& p_\theta(\mathbf{x}^{\mathbf{c}}_{t-1}|\mathbf{x}^{\mathbf{c}}_{t}) = q(\mathbf{x}^{\mathbf{c}}_{t-1}|\mathbf{x}^{\mathbf{c}}_{t},\text{NN}_\theta(\mathbf{x}^{\mathbf{c}}_{t},t)) \nonumber \\ = & \operatorname{Bern}\left(\left[\text{NN}_\theta(\mathbf{x}^{\mathbf{c}}_{t},t) + (1-\text{NN}_\theta(\mathbf{x}^{\mathbf{c}}_{t},t) )\frac{n(t-1)-1}{n(t)-1}\right]\odot\mathbf{x}^{\mathbf{c}}_{t}\right)
\end{align}
Here $\text{NN}_\theta(\mathbf{x}^{\mathbf{c}}_{t},t)$ refers to a probability distribution over $K$-dim, \emph{e.g.}, outputs after $\operatorname{softmax}$. Each parameter in $p_\theta\left(\mathbf{x}_{t-1}^{\mathbf{c}} \mid \mathbf{x}_t^{\mathbf{c}}\right)$ can be viewed as an interpolation between the constant value $\frac{n(t-1)-1}{n(t)-1}$ and 1, weighted by $\mathrm{NN}_\theta(\mathbf{x}^{\mathbf{c}}_{t},t)$. 

Moreover, we propose incorporating the property of the forward process where $\mathbf{x}_{t-1}^{\mathbf{c}}$ is strictly contained within $\mathbf{x}_t^{\mathbf{c}}$, expressed as $\left[\mathbf{x}_t^{\mathbf{c}}\right]^T \mathbf{x}_{t-1}^{\mathbf{c}}=\mathbf{1}^T \mathbf{x}_{t-1}^{\mathbf{c}}$. This property is integrated into parameterization by ensuring that $\mathrm{NN}_\theta\left(\mathbf{x}_t^{\mathbf{c}}, t\right)$ has non-zero values only for categories within $\mathbf{x}_t^{\mathbf{c}}$, satisfying $[\mathrm{NN}_\theta\left(\mathbf{x}_t^{\mathbf{c}}, t\right)]^T\left(\mathbf{1}-\mathbf{x}_t^{\mathbf{c}}\right)=0$. Practically, such condition can be satisfied by ading $-\infty$ to the logits before the $\operatorname{softmax}$ operation. The prior distribution is set as the all-one vector, \emph{i.e.}, $p_\theta(\mathbf{x}^{\mathbf{c}}_T) = \operatorname{Bern}(\mathbf{1})$.

\subsection{Training Procedure} 
\label{sec:training}
We insert the formulation in Eq.~\ref{eq:marginal}, Eq.~\ref{eq:posterior} and Eq.~\ref{eq:predicted} into the Variational lower bound in Eq.~\ref{eq:vlb} to derive the final objective for shortlisting models. We start with the first term $L_T$. As mentioned above, $p_\theta(\mathbf{x}^{\mathbf{c}}_T) = \operatorname{Bern}(\mathbf{1})$. We put the time step $T$ into Eq.~\ref{eq:marginal}, 
\begin{align}
    q(\mathbf{x}^{\mathbf{c}}_T|\mathbf{x}^{\mathbf{c}}_{0}) = \operatorname{Bern}\left(\mathbf{x}^{\mathbf{c}}_{0} + (1-\mathbf{x}^{\mathbf{c}}_{0})\frac{n(T)-1}{K-1}\right) \nonumber  = \operatorname{Bern}\left(\mathbf{x}^{\mathbf{c}}_{0} + (1-\mathbf{x}^{\mathbf{c}}_{0})\frac{K-1}{K-1}\right) = \operatorname{Bern}(\textbf{1})
\end{align}
The first term $L_T$ in Eq.~\ref{eq:vlb} becomes:
$L_T 
     =  \mathbb{E}_{q\left(\boldsymbol{x}^{\mathbf{c}}_0\right)} D_{\mathrm{KL}}\left[ \operatorname{Bern}(\mathbf{1}) \| \operatorname{Bern}(\mathbf{1})\right] = 0$. 
For the last term $L_0$, with $n(0)=1$ the $p_\theta(\mathbf{x}^{\mathbf{c}}_0|\mathbf{x}^{\mathbf{c}}_1)$ in Eq.~\ref{eq:predicted} can be expressed as $p_\theta(\mathbf{x}^{\mathbf{c}}_0|\mathbf{x}^{\mathbf{c}}_1)=\operatorname{Bern}(\text{NN}_\theta(\mathbf{x}^{\mathbf{c}}_{t},t))$. Then $L_0$ is expressed as: 
\begin{align}
    L_0 &= -\mathbb{E}_{q\left(\boldsymbol{x}^{\mathbf{c}}_1 \mid \boldsymbol{x}^{\mathbf{c}}_0\right)}\left[\log p_\theta\left(\boldsymbol{x}^{\mathbf{c}}_0 \mid \boldsymbol{x}^{\mathbf{c}}_1\right)\right] \nonumber \\
    &= -\mathbb{E}_{q\left(\boldsymbol{x}^{\mathbf{c}}_1 \mid \boldsymbol{x}^{\mathbf{c}}_0\right)} [\log \left\langle\text{NN}_\theta(\mathbf{x}^{\mathbf{c}}_{1},t), \boldsymbol{x}^{\mathbf{c}}_0\right\rangle+ 
    \begin{cases}\log \left\langle 1-\text{NN}_\theta(\mathbf{x}^{\mathbf{c}}_{1},t), \boldsymbol{x}^{\mathbf{c}}_1-\boldsymbol{x}^{\mathbf{c}}_0\right\rangle,&\|\mathbf{x}^{\mathbf{c}}_{1}-\mathbf{x}^{\mathbf{c}}_{0}\| >0 \\ 0,&\|\mathbf{x}^{\mathbf{c}}_{1}-\mathbf{x}^{\mathbf{c}}_{0}\| =0\end{cases} \nonumber
\end{align}
Here $\langle \cdot \rangle$ denotes the inner product. Next, we focus on the term $L_{t-1}$, and for simplicity we use $\text{pred}_\theta(\mathbf{x}^{\mathbf{c}}_t)$ as a shorted notation for $[\text{NN}_\theta(\mathbf{x}^{\mathbf{c}}_{t},t) + (1-\text{NN}_\theta(\mathbf{x}^{\mathbf{c}}_{t},t) )\frac{n(t-1)-1}{n(t)-1}]\odot\mathbf{x}^{\mathbf{c}}_{t})$, correspondingly,  $\text{gd}(\mathbf{x}^{\mathbf{c}}_t)$ for $\mathbf{x}^{\mathbf{c}}_{0} + [(1-\mathbf{x}^{\mathbf{c}}_{0})\odot\mathbf{x}^{\mathbf{c}}_{t}] \frac{n(t-1)-1}{n(t)-1}$. 
\begin{align}
\label{eq:lt}
    L_{t-1} 
    = \mathbb{E}_{q\left(\boldsymbol{x}^\mathbf{c}_t \mid \boldsymbol{x}^\mathbf{c}_0 \right)}\left[D_{\mathrm{KL}}\left[\operatorname{Bern}(\text{gd}(\mathbf{x}^{\mathbf{c}}_t)) \| \operatorname{Bern}(\text{pred}_\theta(\mathbf{x}^{\mathbf{c}}_t))\right]\right]
\end{align}
The KL divergence between the Multivariate Bernoulli distribution is extended as:
\begin{align}
\label{eq:multi_vari}
     &D_{\mathrm{KL}}[\operatorname{Bern}(\text{gd}(\mathbf{x}^{\mathbf{c}}_t)) \| \operatorname{Bern}(\text{pred}_\theta(\mathbf{x}^{\mathbf{c}}_t))] \nonumber \\&=\sum_{i, (\mathbf{x}^{\mathbf{c}}_t)^i>0} \left(\text{gd}^i (\mathbf{x}^{\mathbf{c}}_t)\log \frac{\text{gd}^i(\mathbf{x}^{\mathbf{c}}_t)}{\text{pred}^i_\theta(\mathbf{x}^{\mathbf{c}}_t)} 
    + (1-\text{gd}^i(\mathbf{x}^{\mathbf{c}}_t)) \log \frac{1-\text{gd}^i(\mathbf{x}^{\mathbf{c}}_t)}{1-\text{pred}^i_\theta(\mathbf{x}^{\mathbf{c}}_t)}\right) 
\end{align}

Here, we use the superscript $i$ to denote the $i$-th dimension.

\subsection{Mitigating Gradient Vanishing.} 

We observe that directly optimizing the KL divergence of a multi-dimensional Bernoulli distribution, as in Eq.~\ref{eq:lt}, can lead to optimization failure, with the process stalled from the beginning. This issue is likely due to gradient vanishing, where the gradients become too small to drive effective parameter updates. We formally investigate this issue in the following. 

Taking dimension $i$ in the $K$ dimensions and $(\mathbf{x}^{\mathbf{c}}_t)^i>0$ as an example, the gradient towards the parameter $\theta$ is 
$\nabla _ \theta D_{\mathrm{KL}}[\operatorname{Bern}(\text{gd}^i(\mathbf{x}^{\mathbf{c}}_t) \| \operatorname{Bern}(\text{pred}^i_\theta(\mathbf{x}^{\mathbf{c}}_t)] 
    = -(\frac{\text{gd}^i(\mathbf{x}^{\mathbf{c}}_t)}{\text{pred}^i_\theta(\mathbf{x}^{\mathbf{c}}_t)}-\frac{1-\text{gd}^i(\mathbf{x}^{\mathbf{c}}_t)}{1-\text{pred}^i_\theta(\mathbf{x}^{\mathbf{c}}_t)})\nabla_\theta \text{pred}^i_\theta(\mathbf{x}^{\mathbf{c}}_t)$. 
And we denote the above gradient term as $\nabla_\theta D^i_{\mathrm{KL}}$ for simplicity in the following. 
Recall that $\text{gd}^i(\mathbf{x}^{\mathbf{c}}_t)$ and $\text{pred}^i_\theta(\mathbf{x}^{\mathbf{c}}_t)$ are both interpolations between $1$ and $\frac{n(t-1)-1}{n(t)-1}$ as discussed in Section~\ref{sec:reverse}, we have: $ \frac{n(t-1)-1}{n(t)-1} \leq \text{gd}^i(\mathbf{x}^{\mathbf{c}}_t),\text{pred}^i_\theta(\mathbf{x}^{\mathbf{c}}_t) \leq 1$. 
Consider the common situation when $(\mathbf{x}^\mathbf{c}_0)^i = 1$, and the network prediction $\text{NN}^i_\theta(\mathbf{x}^\mathbf{c}_t,t)$ holds a non-zero value. The norm of weight satisfies that:
$\left\| \frac{\text{gd}^i(\mathbf{x}^{\mathbf{c}}_t)}{\text{pred}_\theta(\mathbf{x}^{\mathbf{c}}_t)(i)} - \frac{1-\text{gd}^i(\mathbf{x}^{\mathbf{c}}_t)}{1-\text{pred}^i_\theta(\mathbf{x}^{\mathbf{c}}_t)} \right\|_2 \leq  \frac{n(t)-1}{n(t-1)-1}$.
Combining with $\nabla_\theta \text{pred}^i_\theta(\mathbf{x}^{\mathbf{c}}_t) 
= \frac{n(t)-n(t-1)}{n(t)-1} \nabla_\theta \text{NN}^i_\theta(\mathbf{x}^{\mathbf{c}}_{t},t)$, we could obtain the following bounded condition over the gradient norm of $\nabla_\theta D^i_{\mathrm{KL}}$:
\begin{align}
\label{eq:norm_bound}
    &\|\nabla_\theta D^i_{\mathrm{KL}} \|_2
    =\left\|\frac{\text{gd}^i(\mathbf{x}^{\mathbf{c}}_t)}{\text{pred}^i_\theta(\mathbf{x}^{\mathbf{c}}_t)}-\frac{1-\text{gd}^i(\mathbf{x}^{\mathbf{c}}_t)}{1-\text{pred}^i_\theta(\mathbf{x}^{\mathbf{c}}_t)}\right\|_2 \left\|\nabla_\theta \text{pred}^i_\theta(\mathbf{x}^{\mathbf{c}}_t)\right\|_2 \nonumber \\
    &\leq \frac{n(t)-1}{n(t-1)-1} \frac{n(t)-n(t-1)}{n(t)-1} \left\|\nabla_\theta \text{NN}^i_\theta(\mathbf{x}^{\mathbf{c}}_{t},t)\right\|_2 =  \frac{n(t)-n(t-1)}{n(t-1)-1} \left\| \nabla_\theta \text{NN}^i_\theta(\mathbf{x}^{\mathbf{c}}_{t},t) \right\|_2
\end{align}

Note that the bounds involve the gradient term $\nabla_\theta \mathrm{NN}^i_\theta(\mathbf{x}^{\mathbf{c}}_{t}, t)$, which is computed directly from the $\operatorname{softmax}$ outputs (without applying the logarithm). However, taking gradients through the $\operatorname{softmax}$ function directly often leads to vanishing gradients, particularly in high-dimensional settings where the outputs $\mathrm{NN}^i_\theta(\mathbf{x}^{\mathbf{c}}_{t}, t)$ can be very small initially. We additionally provide formal illustration in Appendix~\ref{app:grad_vanish}.

To mitigate this issue, we propose scaling the gradient in Eq.~\ref{eq:norm_bound} by $\frac{1}{\mathrm{NN}^i_\theta(\mathbf{x}^{\mathbf{c}}_{t}, t)}$, consistent with the widely adopted log-$\operatorname{softmax}$ optimization. Surprisingly, this corresponds directly to the following simplified objective:
\begin{align}
\label{eq:weighted}
    L^{\text{weight}}_{t-1} =  -\mathbb{E}_{q\left(\boldsymbol{x}^\mathbf{c}_t \mid \boldsymbol{x}^\mathbf{c}_0 \right)}\left[\frac{n(t)-n(t-1)}{n(t)-1} \langle  \log \text{NN}_\theta(\boldsymbol{x}^\mathbf{c}_t,t), \boldsymbol{x}^\mathbf{c}_0  \rangle
    \right]
\end{align}
Though derived with heuristic intuition, we formally show in Appendix~\ref{app:derivation_weight} that this reweighted loss (Eq.~\ref{eq:weighted}) can be interpreted as a reasonable approximation.

Moreover, we can derive an even simpler training objective by removing the weight, analogous to the simplified loss used in~\citep{ho2020denoising}, which may provide different practical properties:
\begin{align}
\label{eq:simple}
    L^{\text{simple}}_{t-1} =  -\mathbb{E}_{q\left(\boldsymbol{x}^\mathbf{c}_t \mid \boldsymbol{x}^\mathbf{c}_0 \right)}\left[\langle \log \text{NN}_\theta(\boldsymbol{x}^\mathbf{c}_t,t), \boldsymbol{x}^\mathbf{c}_0  \rangle
    \right]
\end{align}
The above objective is essentially the Cross-Entropy loss between the network prediction and the original data sample.  Unless otherwise specified, we use $L^{\text{simple}}$ for experiments on OpenWebText and $L^{\text{weight}}$ for all other experiments. The likelihood(ELBO) is strictly evaluated using the original ELBO defined in Eq.~\ref{eq:lt}.

\subsection{Candidate Set Size Scheduling.} 
Another important component of the framework is the scheduling function over the size of candidate set, \emph{i.e.} $n(t)$. It is noteworthy that $n(t)$ is not restricted to integer values; rather, it can take any real value within the interval $[1, K]$. We take a similar intuition from \citep{graves2023bayesian}, by considering the normalized vector $\frac{\mathbf{x}^{\mathbf{c}}_t}{\sum_{i=1}^K(\mathbf{x}^{\mathbf{c}}_t)^i}$ as the probability of distribution over vocabulary, then we expect the entropy of the distribution to increase linearly from $t=1$ to $t=T$. Note the expected ones of $\mathbf{x}^{\mathbf{c}}_t$ is exactly $n(t)$,  hence the corresponding entropy of the aforementioned distribution is $\log n(t)$.  Then we can design scheduling function as: $n(t) =  e ^{(\log K) \frac{t}{T}} 
$.

\subsection{Sampling Process} 
The sampling process of shortlisting models can be directly conducted with ancestral sampling based on the learned $p_\theta(\mathbf{x}^{\mathbf{c}}_{t-1}|\mathbf{x}^{\mathbf{c}}_{t})$ with $\mathbf{x}^{\mathbf{c}}_{T} \sim \operatorname{Bern}(\mathbf{1})$ as the starting point. The full pseudocode for training and sampling is provided in Appendix \ref{app:training_sampling_algorithms}.
To ensure the candidate set always contains at least one candidate (i.e., $\mathbf{x}^{\mathbf{c}}_{t} \neq \mathbf{0}$), we empirically set the dimension with the largest Bernoulli parameter to 1 when the sampled vector is a zero vector.

\begin{figure*}[t]
\begin{center}
\includegraphics[width=0.99\textwidth]{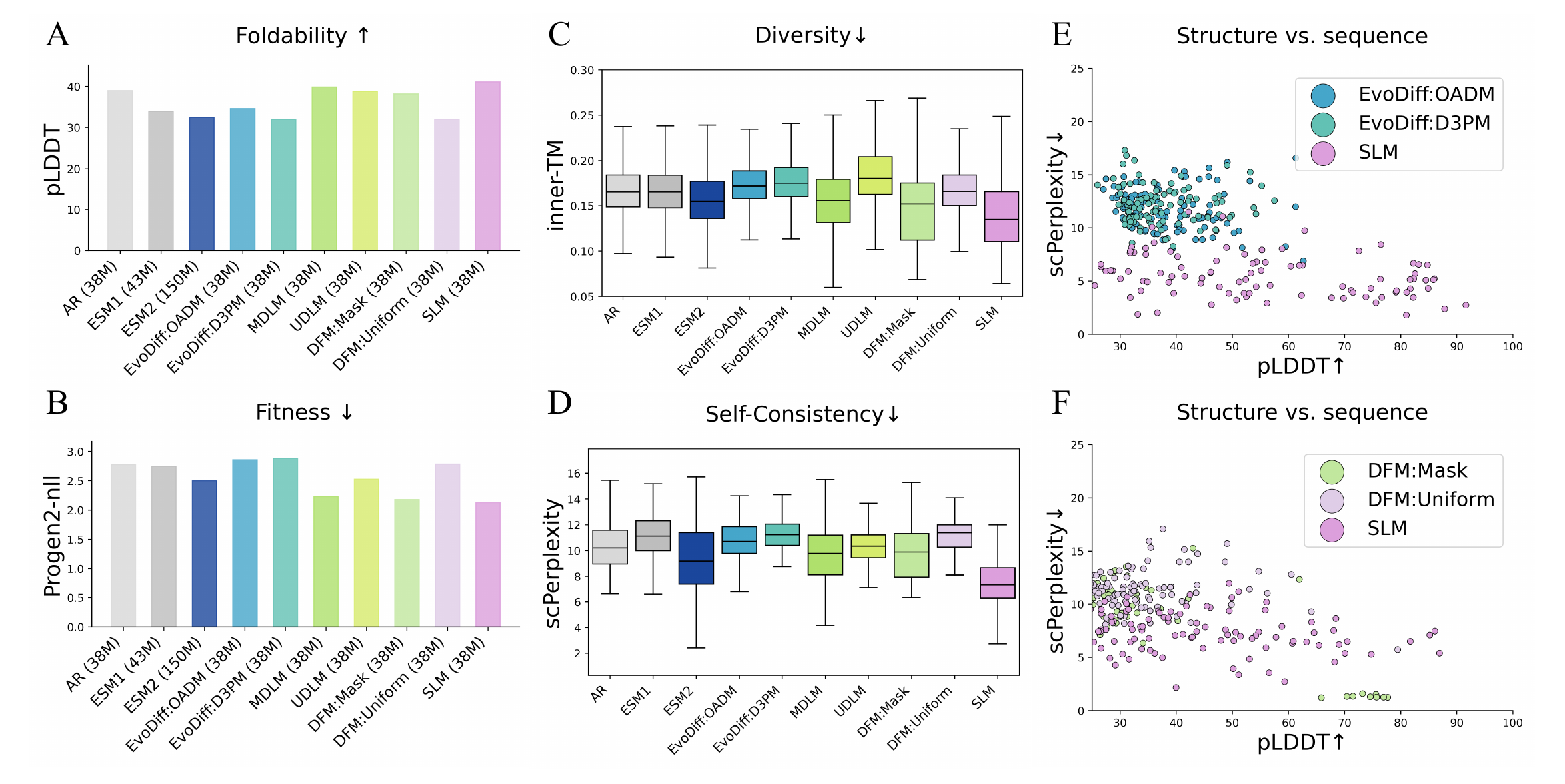}
\end{center}
\caption{\textbf{Quantitative Performance on Protein Sequence Design \methodsy compared to baselines}:  (A-D) pLDDT(A), Progen2-nll(B), scPerplexity(C) and inner-TM(D) scores for sequence sampled from ESM1-43M, ESM2-150M, EvoDiff-OADM-38M, EvoDiff-D3PM-38M, MDLM-38M, DFM-Mask-38M, DFM-Uniform-38M and \method-38M models. (E-F) The joint distribution of pLDDT and scPerplexity from \methodsy model and Diffusion Models(E) and Discrete Flow Matching Models(F).
 }
\label{fig:protein_main}
\end{figure*}

\begin{figure*}[t]
\begin{center}
\includegraphics[width=0.99\textwidth]{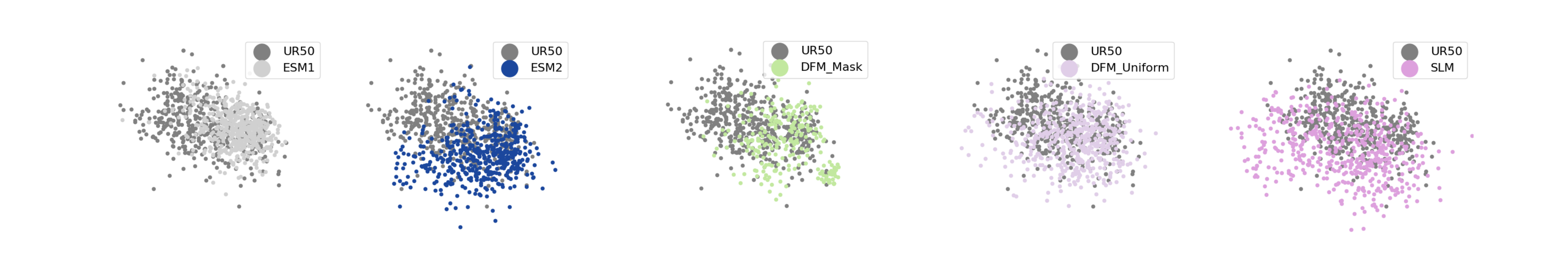}
\caption{SLM not only fits the reference distribution well but also explores a broader outer area under ProstT5 embedding.UR50 for samples from Uniref50 dataset.}
\label{fig:protein_vis}
\end{center}
\end{figure*}

\subsection{Classifier-free Guidance.} 
We show that this simplified formulation offers flexibility to implement classifier-free guidance with an extra class-conditioned shortlist model. Denoting the output of the unconditional model at timestep $t$ as $\text{NN}_\theta(\mathbf{x}^\mathbf{c}_t,t,K)$ and the conditional model as $\text{NN}_\theta(\mathbf{x}^\mathbf{c}_t,t,\text{cls})$. Here $\text{cls} \in [0,K-1] \cap \mathbb{Z}$ denotes the class label. The reverse process based on classifier-free guidance can be obtained as：
$p_\theta^{\text{CFG}}(\mathbf{x}^{\mathbf{c}}_{t-1}|\mathbf{x}^{\mathbf{c}}_{t}) = \operatorname{Bern}\left(\left[\hat{\text{NN}_\theta}(\mathbf{x}^{\mathbf{c}}_{t},t) + (1-\hat{\text{NN}_\theta}(\mathbf{x}^{\mathbf{c}}_{t},t) )\frac{n(t-1)-1}{n(t)-1}\right]\odot\mathbf{x}^{\mathbf{c}}_{t}\right)$
Here the $\hat{\text{NN}_\theta}$ is as:
$\hat{\text{NN}_\theta}(\mathbf{x}^{\mathbf{c}}_{t},t) = \gamma \text{NN}_\theta(\mathbf{x}^\mathbf{c}_t,t,\text{cls})+(1-\gamma) \text{NN}_\theta(\mathbf{x}^\mathbf{c}_t,t,K)$. 
When $\gamma>1$, there can be negative number in $\hat{\text{NN}_\theta}$. Following \citep{stark2024dirichlet}, we project the value of $\hat{\text{NN}_\theta}$ back to the simplex based on \citep{wang2013projection}.

\section{Experiments}

\label{sec:experiments}
In this section, we evaluate our shortlisting models across various discrete data generation tasks and benchmarks. These tasks include language modeling and biological sequence design, the latter of which is especially well suited for non-autoregressive models and for demonstrating the potential of our proposed method.

\subsection{Language Modeling}

\textbf{Text8}: Firstly, we conduct experiments on the text8 dataset~\citep{mahoney2011large} with vocab size of 27. Bits-per-character(BPC) is reported based on the Equation.~\ref{eq:multi_vari}. Details on the dataset can be found in Appendix.\ref{app:dataset_text8}. The results are reported in Table.\ref{tab:text8}, and additional generated samples are provided in Table.\ref{tab:test_last}.

\begin{wraptable}{r}{0.5\textwidth} 
  \centering
    \caption{Bits Per Character (BPC) on Text8 Test Set.}
    \label{tab:text8}
    \small 
    \begin{tabular}{@{} l l c @{}}
        \toprule
        \textbf{Category} & \textbf{Method} & \textbf{BPC ($\downarrow$)} \\ 
        \midrule
        \multirow{3}{*}{\textit{Autoregressive}} 
            & Transformer AR      & \textbf{1.23} \\
            & AR Argmax Flow      & 1.39        \\
            & AR Discrete Flow    & \textbf{1.23} \\
        \midrule
        \multirow{2}{*}{\makecell[l]{\textit{Any-order} \\ \textit{Autoregressive}}} 
            & ARDM                & $\leq$ 1.43  \\
            & MAC                 & $\leq$ 1.40  \\
        \midrule
        \multirow{1}{*}{\textit{Continuous Diffusion}} 
            & Plaid               & $\leq$ 1.48  \\
        \midrule
        \multirow{7}{*}{\textit{Discrete Diffusion}} 
            & Mult. Diffusion     & $\leq$ 1.72  \\
            & D3PM Uniform        & $\leq$ 1.61  \\
            & UDLM                & $\leq$ 1.59  \\
            & D3PM Absorb         & $\leq$ 1.45  \\ 
            & SEDD Absorb         & $\leq$ 1.41  \\ 
            & MDLM                & $\leq$ 1.39  \\ 
            & MD4                 & $\leq$ 1.37  \\
        \midrule 
         \multirow{5}{*}{\textit{Simplex Approaches}} 
             & BFN                 & $\leq$ 1.41  \\
             & DFM                 & $\leq$ 1.41  \\
             & SFM                 &  1.39  \\
         & \cellcolor{mygray}  \textbf{\method}($L^{\text{simple}}$) & \cellcolor{mygray} $\leq$ 1.42 \\
         & \cellcolor{mygray} \textbf{\method}($L^{\text{weight}}$) & \cellcolor{mygray} $\leq$  1.38 \\
        \bottomrule
    \end{tabular}
    \vspace{-20pt}
\end{wraptable}

We compare our shortlisting model with baselines across various categories: $(1)$ autoregressive models: Transformer AR\citep{Vaswani2017Attention}, AR Argmax Flow\citep{hoogeboom2021argmax}, AR Discrete Flow\citep{tran2019discrete}; $(2)$ any-order autoregressive models: ARDM\citep{hoogeboom2021autoregressive}, MAC~\citep{shih2022training}; $(3)$ embedding-space continuous diffusion models: Plaid\citep{gulrajani2024likelihood}); $(4)$ advanced discrete diffusion models: SEDD\citep{lou2023discrete}, MDLM\citep{sahoo2024simple}, UDLM\citep{schiff2024simple}, D3PM variants\citep{austin2021structured}; and $(5)$ simplex-based approaches: BFN\citep{graves2023bayesian}, SFM\citep{cheng2024categorical}.

As aforementioned, we report the BPC of both the shortlisting model(\method) trained with the $L^{\text{simple}}$ in Eq.~\ref{eq:simple} and with the $L^{\text{weight}}$ in Eq.~\ref{eq:weighted}. As demonstrated in Table \ref{tab:text8}, even with the simplified objective, the proposed approach achieves competitive performance compared to other non-autoregressive approaches. Moreover, the reweighted formulation further boosts density estimation performance by better aligning with the original ELBO, as discussed in Section~\ref{sec:training}.

\textbf{OpenWebText}: We further investigate the challenges and potential of simplex-based approaches in large vocabulary settings, over the OpenWebText~\citep{Gokaslan2019OpenWeb} dataset with vocab size of 50527. Detailed results and discussions are provided in Appendix~\ref{app:owt} and Table.~\ref{tab:owt}, highlighting that while our SLM still lags behind advanced autoregressive models in density estimation, it achieves competitive generation performance and significantly outperforms existing simplex-based methods.

\subsection{De novo Design of Protein Sequence}
In this experiment, we focus on the core task of unconditional protein design, and examine various protein properties to demonstrate \method's superiority in this task. We present distribution-level visualizations and analyses in Figure~\ref{fig:protein_vis}, which further highlights the effectiveness and biological relevance of \method. Details of the training and visualization procedures can be found in Appendix~\ref{protein:train} and Appendix~\ref{protein:visual}, respectively.

\textbf{Baselines}: We compare \methodsy against three groups of existing methods: $(1)$ Autoregressive models (AR); $(1)$ Masked language models (MLMs), specifically \textbf{ESM1}\citep{rives2019biological} and \textbf{ESM2}\citep{lin2022language}; $(2)$ Discrete Diffusion Models, represented by two versions of EvoDiff\citep{alamdari2023protein}: \textbf{EvoDiff-OADM},\textbf{EvoDiff-D3PM}, \textbf{MDLM}\citep{sahoo2024simple} and \textbf{UDLM}\citep{schiff2024simple}; $(3)$ Discrete Flow Matching Models\citep{gat2024discrete}: \textbf{DFM-Mask} and \textbf{DFM-Uniform}. Further information of these baselines in Appendix \ref{app:prot_baseline}.

To demonstrate \method's effectiveness in protein sequence generation, we evaluate four key properties: $(1)$~\textbf{Foldability}: structural plausibility predicted by ESMFold \citep{lin2022language}; $(2)$~\textbf{Fitness}: scores predicted by ProGen2-xlarge \citep{nijkamp2023progen2}; $(3)$~\textbf{Self-Consistency}: alignment between sequences folded by ESMFold and inverse-folded by ESM-IF \citep{hsu2022learning}; and $(4)$~\textbf{Diversity}: pairwise inner-TM scores among generated samples. Detailed metric definitions are provided in Appendix~\ref{protein:eval}. As shown in Figure~\ref{fig:protein_main}, \methodsy surpasses all baselines across all metrics and achieves competitive performance compared to ESM2-150M \citep{lin2022language}, demonstrating strong generalization and robustness under restricted vocabularies and complex data distributions.

\subsection{Design of DNA Sequence}

In this part, we focus on the roles of promoters and enhancers, and evaluate \methodsy in this context. Following prior work, we set the language model to use 500 NFE for enhancers and 1000 NFE for promoters. For other models, Non-CFG models use 100 NFE, and CFG variants use 200 NFE.

\subsubsection{Promoter DNA Sequence Design}

We follow the setting in previous work DDSM \citep{avdeyev2023dirichlet} to generate DNA promoter sequences conditioned on the promoter profile.

\begin{wrapfigure}{r}{0.6\textwidth}
  \begin{center}
    \includegraphics[width=0.9\textwidth]{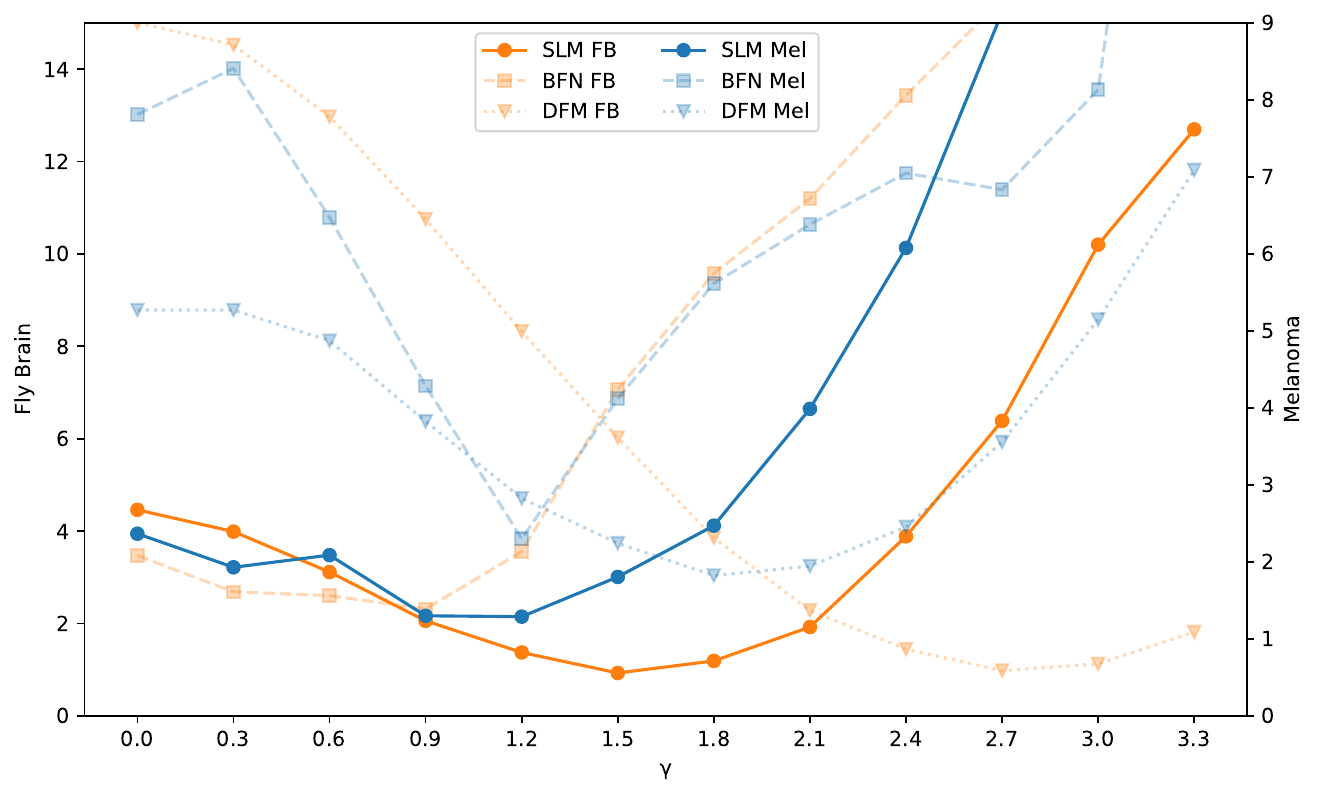}
  \end{center}
  \caption{Performance of SLM under different CFG factor $\gamma$ for unconditional enhancer design.}
  \label{fig:FBD-gamma}
\end{wrapfigure}

\textbf{Data:} We use a dataset of 100{,}000 human promoter sequences, each 1024 base-pairs long, extracted from the Human Promoter Database \citep{hon2017atlas}. Each sequence is paired with a CAGE signal indicating transcriptional activity at each position \citep{shiraki2003cap, forrest2014promoter}. Sequences from chromosomes 8 and 9 form the test set, and the remainder are used for training.

\textbf{Baselines:} We compared the \methodsy method with: $(1)$  flow matching methods including Dirichlet FM \citep{stark2024dirichlet} and Fisher-Flow \citep{davis2024fisher}; $(2)$ autoregressive language model that generates base pairs; $(3)$ Bayesian Flow Networks (BFN) \citep{graves2023bayesian}; and $(4)$ other discrete diffusion methods including two implementations of Bit Diffusion \citep{chen2022analog}, D3PM \citep{austin2021structured}, MDLM~\citep{sahoo2024simple}, UDLM~\citep{schiff2024simple} and simplex-based DDSM \citep{avdeyev2023dirichlet}.

\textbf{Results:} The regulatory activity of the sequences is given by Sei, a model that predicts the regulatory potential of the sequences \citep{chen2022sequence}. We report the mean and standard deviation of MSE between the generated sequences and the target. Our MSE values were measured under the same Sei model as in previous works. As shown in Table.\ref{tab:promoter}, our \methodsy method achieves the lowest MSE, with a smaller standard deviation as well.

{\small
\begin{table*}
\begin{floatrow}
\capbtabbox{
\begin{tabular}{lc}
\toprule
Model & MSE($\downarrow$)\\
\midrule
Autoregressive                   & 0.034$\pm$0.001 \\
Bit Diffusion (bit enc)     & 0.041 \\
Bit Diffusion (one-hot) & 0.040 \\
D3PM-uniform                     & 0.038 \\
UDLM                     & 0.030$\pm$0.001 \\
MDLM                             & 0.028$\pm$0.001 \\
Dirichlet FM                     & 0.034$\pm$0.004 \\
Fisher-Flow                      & 0.029$\pm$0.001 \\
BFN                              & 0.0405$\pm$0.0003 \\
DDSM                             & 0.033 \\
\midrule
\rowcolor{mygray}\method   & \textbf{0.0265$\pm$0.0006} \\
\bottomrule
\end{tabular}
}{
 \caption{Conditional generation over promoter design. BFN results are from our experiments and the other baselines from \citep{davis2024fisher}.}
 \label{tab:promoter}
 \small
}
\capbtabbox{
\begin{tabular}{l@{\hspace{-2pt}}cc}
\toprule
Model & Mel FBD($\downarrow$) & FB FBD($\downarrow$)\\
\midrule
Random           & 619.0$\pm$0.8 & 832.4$\pm$0.3 \\
Autoregressive   & 35.4$\pm$0.5  & 25.7$\pm$1.0  \\
Fisher-Flow      & 27.5$\pm$2.6  & 3.8$\pm$0.3   \\
Dirichlet FM     & 5.3$\pm$0.5   & 15.1$\pm$0.4  \\
BFN              & 3.3$\pm$0.1   & 10.8$\pm$0.6  \\
\rowcolor{mygray} \method              & \textbf{2.2$\pm$0.1}   & \textbf{4.4$\pm$0.2}   \\ 
\midrule
BFN CFG          & 2.3$\pm$0.1   & 2.3$\pm$0.2   \\
Dirichlet FM CFG & 1.9$\pm$0.4   & 1.0$\pm$0.3   \\
\rowcolor{mygray} \method ~CFG          & \textbf{1.4$\pm$0.1}   & \textbf{1.0$\pm$0.1}   \\
\bottomrule
\end{tabular}
}{
 \caption{FBD metric for sequence generation under two datasets. CFG refers to Classifier-Free Guidance.}
 \label{tab:enhancer}
 \small
}
\end{floatrow}
\end{table*}
}

\subsubsection{Enhancer DNA Sequence Design}
We also evaluate \methodsy for enhancer generation, following DirichletFM~\citep{stark2024dirichlet}.

\textbf{Data:} We use 104k enhancer sequences from fly brain cells and 89k from human melanoma cells \citep{janssens2022decoding, atak2021interpretation}, each with a length of 500. Cell type labels were determined by ATAC-seq data\citep{buenrostro2013transposition}, with fly brain cells divided into 81 classes and human melanoma cells into 47 classes based on cell types.

\textbf{Baselines:} In addition to their standard implementations, the baseline models also incorporate classifier-free guidance. We select the optimal classifier-free guidance factor $\gamma$ for all models. The performance of our method under different classifier-free guidance factors $\gamma$ is presented in Figure.\ref{fig:FBD-gamma}. Specific experimental settings and details can be found in Appendix \ref{appendix:CFG}.

\textbf{Results:} We evaluate generated sequences using the Fréchet Biological Distance (FBD) from Dirichlet FM \citep{stark2024dirichlet}, which treats classifier-hidden representations as sequence embeddings and computes FBD as the Wasserstein distance between them. \methodsy achieves the best performance without label guidance and further improves with label guidance (see Table~\ref{tab:enhancer}).

\subsection{Ablation Study on Reweighted Loss}
We conduct ablation experiments comparing the simple loss and the reweighted loss on both enhancer and promoter tasks. Table~\ref{tab:ablation} shows that the reweighted loss achieves superior density estimation, while the simple loss can generate better samples.

{\small
\begin{table}[!t]
\centering
\caption{Ablation study on loss function conducted on DNA sequence design.}
\label{tab:ablation}
\begin{tabular}{lcccccc}
\toprule
\multirow{2}{*}{Loss type} & \multicolumn{2}{c}{Mel} & \multicolumn{2}{c}{FB} & \multicolumn{2}{c}{promoter} \\
\cmidrule(lr){2-3} \cmidrule(lr){4-5} \cmidrule(lr){6-7}
& FBD($\downarrow$) & PPL($\downarrow$) & FBD($\downarrow$) & PPL($\downarrow$) & MSE($\downarrow$) & PPL($\downarrow$) \\
\midrule
$L^{\text{simple}}$ & \textbf{2.1788} & 3.4102 & \textbf{4.4450} & \textbf{3.4618} & 0.0265 & 2.8084 \\
$L^{\text{weight}}$ & 2.5848 & \textbf{3.4018} & 4.9670 & 3.4654 & \textbf{0.0260} & \textbf{2.7672} \\
\bottomrule
\end{tabular}
\end{table}
}

    \section{Comparison to Existing Works}

Our SLM connects closely to several existing approaches. When $K=2$, SLM resembles Bernoulli diffusion~\citep{sohl2015deep}, though it operates within a three-state space $[0,1], [1,0], [1,1]$ rather than Bernoulli diffusion's two-state $(0,1)$ space. The progressive candidate exclusion in SLM also shares conceptual similarities with the SUBS parameterization employed in MDLM~\citep{shi2024simplified,sahoo2024simple}. Unlike mask-based discrete diffusion methods, however, SLM directly operates on the simplex, refining information gradually over time.

Recent simplex-based methods such as DirichletFM~\citep{stark2024dirichlet} define trajectories over the continuous full simplex, whereas our SLM explicitly restricts transitions to discrete centroids or their subspaces, thereby reducing degrees of freedom and enhancing efficiency and interpretability. Additionally, while DirichletFM optimizes an MSE-based loss, SLM adopts a simplified cross-entropy objective (Eq.~\ref{eq:weighted}), alleviating gradient vanishing issues associated with  Bernoulli KL losses and leading to improved performance in challenging tasks.

    \section{Conclusion}
\label{sec:conclusion}
In this paper, we introduce the Shortlisting Model (SLM), a novel discrete generative model inspired by progressive candidate pruning. SLM follows a unique generation trajectory by transitioning from the centroids of the simplex space. With impressive performance across various tasks, SLM offers a simple yet effective alternative for discrete generative modeling.

\paragraph{Limitation} This work focuses on simplex-based discrete generative models, with evaluations conducted on text and biological sequence data. Comprehensive studies on other modalities, such as graphs and images, are beyond the scope of this paper and are left for future work. Additionally, further theoretical analysis and engineering improvements are required to scale our approach to large, real-world applications.

    
    \clearpage

    \bibliographystyle{unsrt}
    \bibliography{refs}

    \clearpage
    \beginappendix
    \section{Mathematical  Derivation}
\label{app:proof}
\subsection{Proof of Proposition 3.1}
Since $\mathbf{x}$ is a vector and the elements of the vector are independent, we only consider the position of a fixed index in all the vectors. In the following, all instances of $\mathbf{x}$ are redefined as scalars. First, we prove the following proposition:

\begin{equation}
    p(\mathbf{x}_{t}^{\mathbf{c}} = 1 \mid \mathbf{x}_{0}^{\mathbf{c}} = 0) = \frac{n(t)-1}{K-1}
\end{equation}

When $t=0$, $p(\mathbf{x}_{0} = 1 \mid \mathbf{x}_{0} = 0) = 0$ is obvious. Thus, we can proceed with induction on $t$.

\begin{align}
     & q(\mathbf{x}_{t}^{\mathbf{c}} = 1 \mid \mathbf{x}_{0}^{\mathbf{c}} = 0) \nonumber \\
    = & \frac{n(t-1)-1}{K-1} + \frac{n(t) - n(t-1)}{K-n(t-1)} \left(1-\frac{n(t-1)-1}{K-1}\right) \nonumber \\
    = & \frac{n(t)-1}{K-1}
\end{align}

Since $q(\mathbf{x}_{t} = 1 \mid \mathbf{x}_{0} = 1) = 1$, the two cases can be combined into $q(\mathbf{x}_{t} \mid \mathbf{x}_{0}) = \operatorname{Bern}(\mathbf{x}_{0} + (1-\mathbf{x}_{0})\frac{n(t)-1}{K-1})$, whose vector form is given by Eq.~\ref{eq:marginal}.

The only non-trivial case in the posterior distribution is:

\begin{align}
    & q(\mathbf{x}^{\mathbf{c}}_{t-1} = 1 \mid \mathbf{x}^{\mathbf{c}}_{t} = 1,\mathbf{x}^{\mathbf{c}}_{0} = 0) \nonumber \\
    = & \frac{q(\mathbf{x}^{\mathbf{c}}_{t-1} = 1, \mathbf{x}^{\mathbf{c}}_{t} = 1\mid \mathbf{x}^{\mathbf{c}}_{0} = 0)}{q(\mathbf{x}^{\mathbf{c}}_{t} = 1\mid \mathbf{x}^{\mathbf{c}}_{0} = 0)} \nonumber \\ 
    = & \frac{q(\mathbf{x}^{\mathbf{c}}_{t-1} = 1\mid \mathbf{x}^{\mathbf{c}}_{0} = 0)}{q(\mathbf{x}^{\mathbf{c}}_{t} = 1\mid \mathbf{x}^{\mathbf{c}}_{0} = 0)} \nonumber \\ 
    = & \frac{n(t-1)-1}{n(t)-1}
\end{align}

Only when $\mathbf{x}^{\mathbf{c}}_{0} = 1$, $q(\mathbf{x}^{\mathbf{c}}_{t-1} = 1 \mid \mathbf{x}^{\mathbf{c}}_{t} = 1,\mathbf{x}^{\mathbf{c}}_{0} = 1) = 1$. In all other cases, the probability is 0. Therefore, the result of Eq.~\ref{eq:posterior} can be given.

\subsection{Gradient Vanishing of $\nabla_\theta \text{NN}^i_\theta(x^{\mathbf{c}}_t,t)$}
\label{app:grad_vanish}

Note that $\text{NN}^i_\theta(x^{\mathbf{c}}_t,t)$ is essentially the output of the $\operatorname{softmax}$, which could be further expressed as: 

\begin{align}
    \text{NN}^i_\theta(\mathbf{x}^{\mathbf{c}}_t, t) 
    = \frac{
        \exp\left( f^i_\theta(\mathbf{x}^{\mathbf{c}}_t, t) \right)
    }{
        \sum_{j} \exp\left( f^j_\theta(\mathbf{x}^{\mathbf{c}}_t, t) \right)
    }
\end{align}

where $f_\theta$ denotes the raw output of the neural network. The gradient with respect to the $\operatorname{softmax}$ input is:
\begin{align}
    \frac{\partial \text{NN}^i_\theta(\mathbf{x}^{\mathbf{c}}_t, t)}{\partial f^k_\theta(\mathbf{x}^{\mathbf{c}}_t, t)}
    = \text{NN}^i_\theta(\mathbf{x}^{\mathbf{c}}_t, t) \left( \delta_{ik} - \text{NN}^k_\theta(\mathbf{x}^{\mathbf{c}}_t, t) \right)
\end{align}
The norm is as $\| \text{NN}^i_\theta(\mathbf{x}^{\mathbf{c}}_t, t) \left( 1 - \text{NN}^i_\theta(\mathbf{x}^{\mathbf{c}}_t, t) \right)\|_2$ when $k=i$; and $\|\text{NN}^i_\theta(\mathbf{x}^{\mathbf{c}}_t, t) \text{NN}^k_\theta(\mathbf{x}^{\mathbf{c}}_t, t) \|_2$  when $k\neq i$. Both case the norm is strictly bounded with the $\|\text{NN}^i_\theta(\mathbf{x}^{\mathbf{c}}_t, t)\|_2$. At the initial training stage, $\text{NN}^i_\theta(\mathbf{x}^{\mathbf{c}}_t, t)$ may become uniformly small in high-dimensional settings, leading to vanishing gradients and causing the optimization to stall.

\subsection{Clarification for the relationship between Eq.~\ref{eq:weighted} and ELBO}
We provide a formal derivation of the reweighted loss, which originates from an analysis of the gradient of the KL divergence. Let $\text{NN}_{\theta}^{i}(x^{\mathbf{c}}_t,t)$ denote the model $\operatorname{softmax}$ output at $i$-th dimention. The corresponding predicted Bernoulli distribution's parameter at $i$-th dimention can be expressed as: $\text{NN}_\theta^{i}(\mathbf{x}^{\mathbf{c}}_{t},t) + (1-\text{NN}_\theta^{i}(\mathbf{x}^{\mathbf{c}}_{t},t) )\frac{n(t-1)-1}{n(t)-1}$. For notion simplicity, we define $q = \frac{n(t-1)-1}{n(t)-1}$. Taking the gradient of the KL divergence with respect to $\theta$, we obtain the following expressions:

\begin{itemize}
    \item For the case where $x^{i}_0 = 0$:
        \begin{align}
            & \nabla _ \theta D_{\mathrm{KL}}[\operatorname{Bern}(\text{gd}(\mathbf{x}^{\mathbf{c}}_t)(i)) \| \operatorname{Bern}(\text{pred}_\theta(\mathbf{x}^{\mathbf{c}}_t)(i))] \nonumber \\
            = & -\left[\frac{q}{\text{NN}_{\theta}^{i}(x^{\mathbf{c}}_t,t)+(1-\text{NN}_{\theta}^{i}(x^{\mathbf{c}}_t,t))q}-\frac{1}{1-\text{NN}_{\theta}^{i}(x^{\mathbf{c}}_t,t)}\right] \cdot (1-q) \nabla_\theta \text{NN}_{\theta}^{i}(x^{\mathbf{c}}_t,t)
        \end{align}
    \item For the case where $x^{i}_0 = 1$:
        \begin{align}
            & \nabla _ \theta D_{\mathrm{KL}}[\operatorname{Bern}(\text{gd}(\mathbf{x}^{\mathbf{c}}_t)(i)) \| \operatorname{Bern}(\text{pred}_\theta(\mathbf{x}^{\mathbf{c}}_t)(i))] \nonumber \\
            = & -\frac{1}{\text{NN}_{\theta}^{i}(x^{\mathbf{c}}_t,t)+(1-\text{NN}_{\theta}^{i}(x^{\mathbf{c}}_t,t))q} \cdot (1-q) \nabla_\theta \text{NN}_{\theta}^{i}(x^{\mathbf{c}}_t,t)
        \end{align}
\end{itemize}

We consider the initial stage of training with high-dimensional data, hence the init value of $\text{NN}_{\theta}^{i}(x^{\mathbf{c}}_t,t)$ is relatively small. For the cases when $x^{i}_0 = 0$, the $\text{NN}_{\theta}^{i}(x^{\mathbf{c}}_t,t)$ already approach the optimal value and also the weight of gradient is approximately zero:
\begin{align}
    -\left[\frac{q}{\text{NN}_{\theta}^{i}(x^{\mathbf{c}}_t,t)+(1-\text{NN}_{\theta}^{i}(x^{\mathbf{c}}_t,t))q}-\frac{1}{1-\text{NN}_{\theta}^{i}(x^{\mathbf{c}}_t,t)}\right] \approx -\left[ \frac{q}{q} - 1\right] = 0
\end{align} 
Therefore, we could mainly focus on the case when 
 $x^{i}_0 = 1$. As we mentioned in the above Appendix.~\ref{app:grad_vanish}, the term $\nabla_\theta \text{NN}_{\theta}^{i}(x^{\mathbf{c}}_t,t)$ could have the vanishing issues due to the property of $\operatorname{softmax}$. However, the scale weight $\frac{1}{\text{NN}_{\theta}^{i}(x^{\mathbf{c}}_t,t)+(1-\text{NN}_{\theta}^{i}(x^{\mathbf{c}}_t,t))q}$ is bounded, \emph{i.e.}, $\frac{1}{\text{NN}_{\theta}^{i}(x^{\mathbf{c}}_t,t)+(1-\text{NN}_{\theta}^{i}(x^{\mathbf{c}}_t,t))q} \leq 1$ as $0 \leq \text{NN}_{\theta}^{i}(x^{\mathbf{c}}_t,t),q \leq 1$, and hence could not help enhance the signal. Motivated by the widely optimized $\operatorname{logsoftmax}$ or $\operatorname{logsumexp}$ where the gradient scale weight is as $\frac{1}{\text{NN}_{\theta}^{i}(x^{\mathbf{c}}_t,t)}$, we fix the gradient as:
 \begin{align}
     -\frac{1}{\text{NN}_{\theta}^{i}(x^{\mathbf{c}}_t,t)+(1-\text{NN}_{\theta}^{i}(x^{\mathbf{c}}_t,t))q} \cdot (1-q) \nabla_\theta \text{NN}_{\theta}^{i}(x^{\mathbf{c}}_t,t) \rightarrow &-\frac{1}{\text{NN}_{\theta}^{i}(x^{\mathbf{c}}_t,t)} \cdot (1-q) \nabla_\theta \text{NN}_{\theta}^{i}(x^{\mathbf{c}}_t,t) \nonumber \\
     & = \nabla_\theta L^{\text{weight}}
 \end{align}
Note that the optimization challenge typically arises during the initial training stages, where our proposed objective can provide effective support. Direct optimization of the original ELBO in later training stages or epochs may be possible and could further improve density estimation performance. We leave exploring this direction for future work.

\label{app:derivation_weight}

\begin{figure}[htbp]
\centering
\begin{minipage}[t]{0.48\textwidth}
\centering
\includegraphics[width=0.9\textwidth]{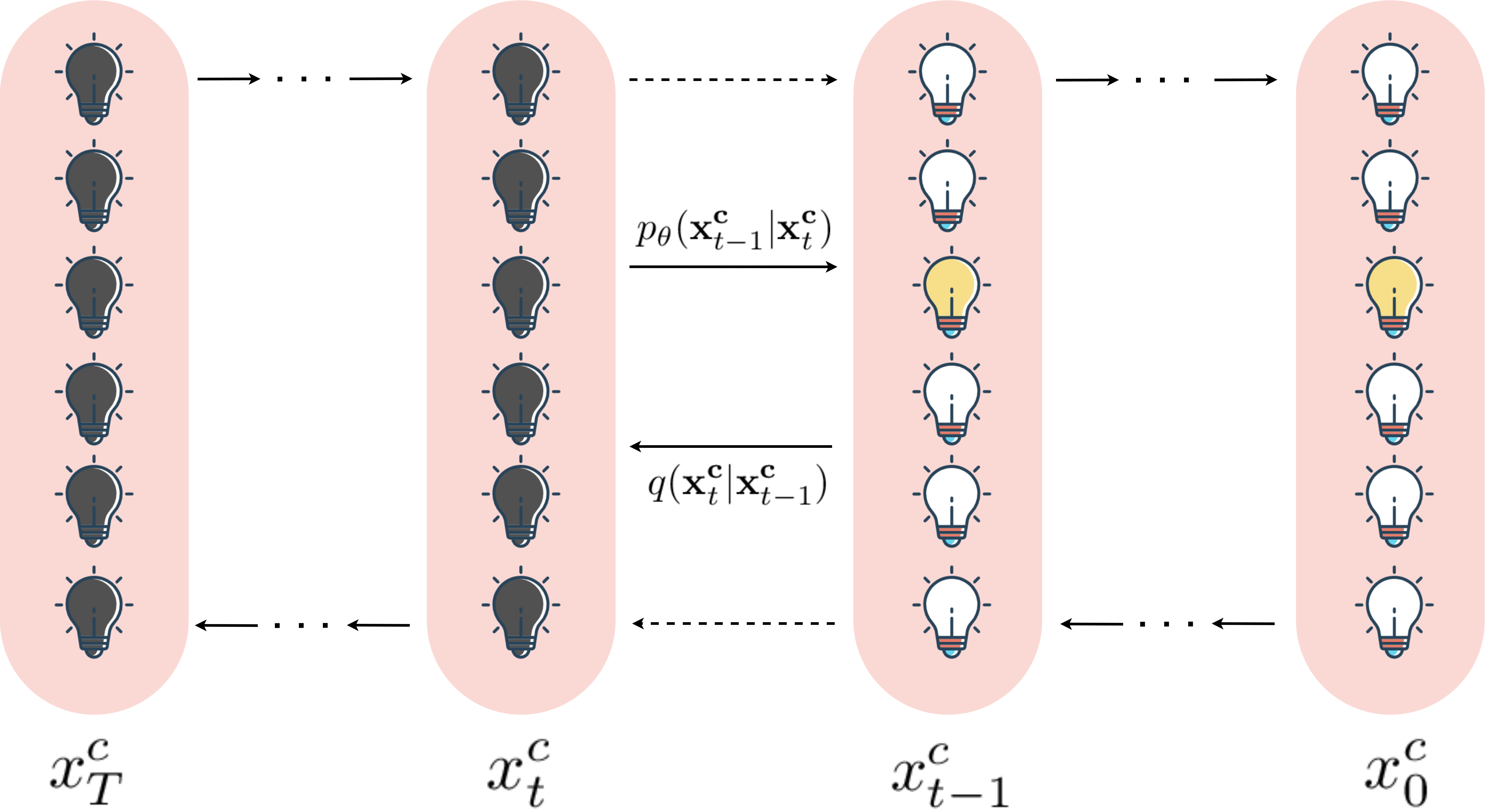}
\end{minipage}
\begin{minipage}[t]{0.48\textwidth}
\centering
\includegraphics[width=0.9\textwidth]{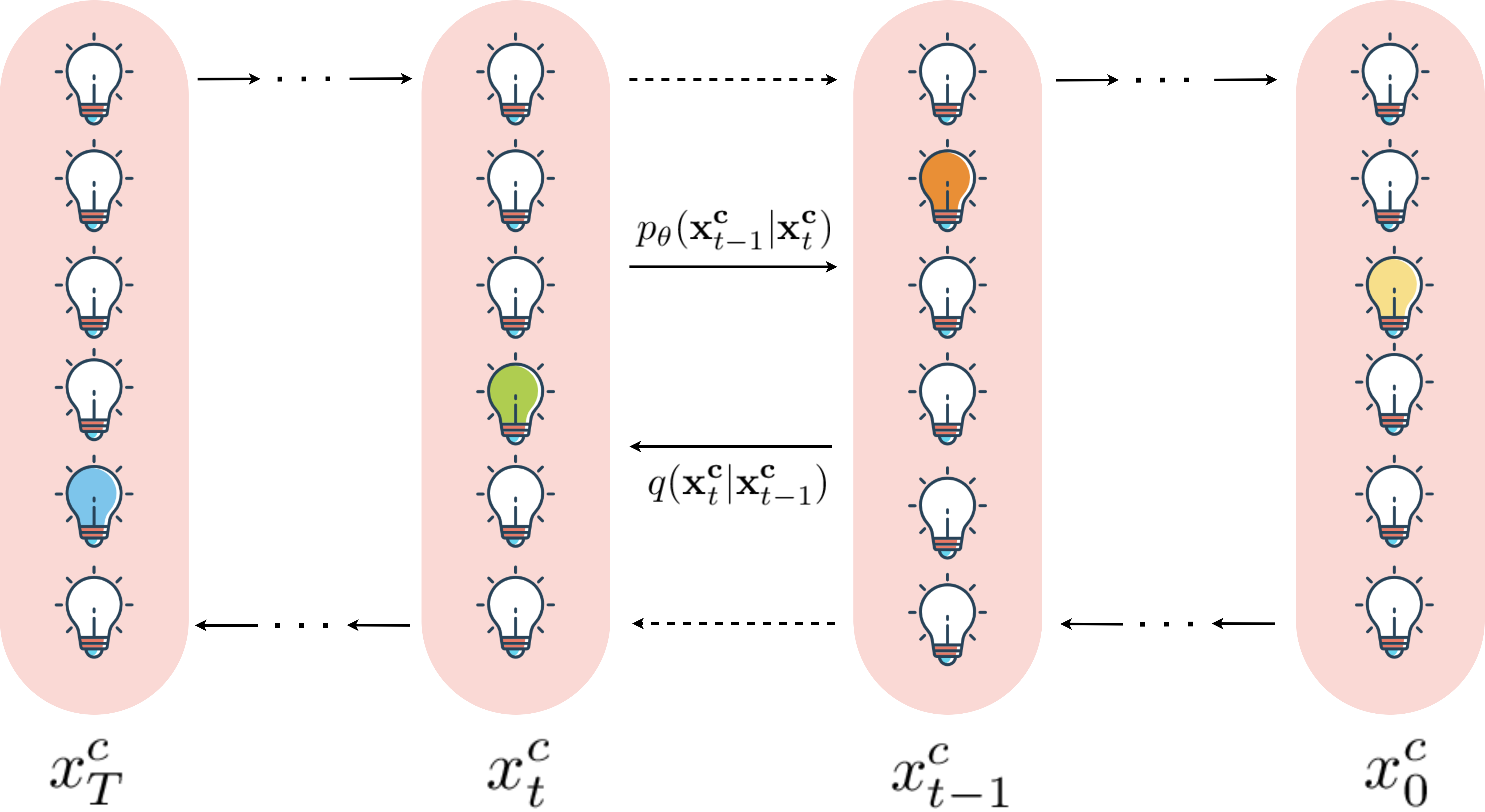}
\end{minipage}
\caption{forward and reverse process of MDM(Left) and D3PM-Uniform(Right)}
\label{fig:overview_mdm_d3pm}
\end{figure}

\section{Algorithms}
\subsection{Visualization of the forward and reverse process of MDLM and D3PM-Uniform}
\label{app:cmp_visual}
In this section, the forward and reverse process of MDLM and D3PM-Uniform are visualized in Figure. \ref{fig:overview_mdm_d3pm}.

\subsection{Training and Sampling Algorithms}
\label{app:training_sampling_algorithms}
In this section, we provide detailed information about the training and sampling processes of \methodsy, with pseudo code as shown in Algorithm.\ref{alg:forward_process}, Algorithm.\ref{alg:training} and Algorithm.\ref{alg:sampling}, with code implementations in PyTorch, as shown in Listing.\ref{code:training} and \ref{code:sampling}.

\begin{algorithm}[h]
  \caption{Forward Process $q(x_t^\mathbf{c} \mid x_0^\mathbf{c})$}
  \label{alg:forward_process}
\begin{algorithmic}
  \STATE {\bfseries Input:} one-hot data $x_0^\mathbf{c}$, time $t$
  \STATE $n(t) \leftarrow e^{(\log K) \frac{t}{T}}$
  \STATE $\text{Bern\_param} = \frac{n(t) - 1}{K - 1}$
  \FOR{$i=0$ {\bfseries to} $K-1$}
        \IF{$x_0^\mathbf{c}[i] == 1$}
            \STATE $x_t^\mathbf{c}[i] \leftarrow 1$
        \ELSE
            \STATE $x_t^\mathbf{c}[i] \leftarrow \text{sample from } \text{Bern\_param} $
        \ENDIF
    \ENDFOR
    \STATE {\bfseries Return} $x_t^\mathbf{c}$
\end{algorithmic}
\end{algorithm}

\begin{algorithm}[h]
  \caption{Training}
  \label{alg:training}
\begin{algorithmic}
  \STATE {\bfseries Input:} one-hot data $x_0^\mathbf{c}$, class label $\text{cls} \in [0, K-1] \cap \mathbb{Z}$
  \STATE Sample $t \sim U(0,1)$
  \STATE $n(t) \leftarrow e^{(\log K) \frac{t}{T}}, n(t-1) \leftarrow e^{(\log K) \frac{t-1}{T}}$
  \STATE $x_t^\mathbf{c} \leftarrow q(x_t^\mathbf{c} \mid x_0^\mathbf{c})$
  \STATE $\text{flag} \sim U(0,1)$
  \IF{$\text{flag} > 0.3$}
        \STATE $\text{cls\_inp} \leftarrow \text{cls}$
  \ELSE
        \STATE $\text{cls\_inp} \leftarrow K$
  \ENDIF
  \STATE $L \leftarrow \log (\langle\text{NN}_\theta(x_t, \text{cls\_inp}, t), x_0^\mathbf{c} \rangle)$
  \STATE {\bfseries Return} $L$
\end{algorithmic}
\end{algorithm}

\begin{algorithm}[t!]
  \caption{Sampling of Shortlisting Model} 
  \label{alg:sampling}
\begin{algorithmic}
  \STATE {\bfseries Input:} class label $\text{cls} \in [0, K-1] \cap \mathbb{Z}$, classifier-free guidance (CFG) factor $\gamma \in \mathbb{R}$ 
  \STATE $x_t^\mathbf{c} \leftarrow \mathbf{1}$
  \FOR{$t=T$ {\bfseries to} $1$}
        \STATE $n(t) \leftarrow e^{(\log K) \frac{t}{T}}, n(t-1) \leftarrow e^{(\log K) \frac{t-1}{T}}$
        \IF{CFG}
            \STATE $\hat{\text{NN}_\theta}(\mathbf{x}^{\mathbf{c}}_{t},t) \leftarrow \gamma \cdot \text{NN}_\theta(x_t^\mathbf{c}, t, \text{cls}) + (1-\gamma) \cdot \text{NN}_\theta(x_t^\mathbf{c}, t, K)$
        \ELSE
            \STATE $\hat{\text{NN}_\theta}(\mathbf{x}^{\mathbf{c}}_{t},t) \leftarrow \text{NN}_\theta(x_t^\mathbf{c}, t)$
        \ENDIF
        \STATE $\text{pred}_\theta \leftarrow \hat{\text{NN}_\theta}(\mathbf{x}^{\mathbf{c}}_{t},t) + \frac{n(t-1) -1}{n(t)-1} (1-\hat{\text{NN}_\theta}(\mathbf{x}^{\mathbf{c}}_{t},t))$
        \STATE $x_t^\mathbf{c} \leftarrow \text{sample from } \text{pred}_\theta$
  \ENDFOR
  \IF{CFG}
        \STATE $\hat{\text{NN}_\theta}(\mathbf{x}^{\mathbf{c}}_{t},0) \leftarrow \gamma \cdot \text{NN}_\theta(x_t^\mathbf{c}, 0, \text{cls}) + (1-\gamma) \cdot \text{NN}_\theta(x_t^\mathbf{c}, 0, K)$
    \ELSE
        \STATE $\hat{\text{NN}_\theta}(\mathbf{x}^{\mathbf{c}}_{t},0) \leftarrow \text{NN}_\theta(x_t^\mathbf{c}, 0)$
    \ENDIF
  \STATE {\bfseries Return} $\arg\max \hat{\text{NN}_\theta}(\mathbf{x}^{\mathbf{c}}_{t},0)$
\end{algorithmic}
\end{algorithm}

\section{Experimental Details}
\subsection{Language Modeling}

\subsubsection{Additional Information for Text8 Experiments}
\label{app:dataset_text8}
The text8 dataset~\citep{mahoney2011large} is a medium-sized character-level corpus with a vocabulary size of 27. It includes 26 lowercase letters and a space token, sourced from the m English Wikipedia dataset by removing punctuation and converting all text to lowercase. The data processing is directly following the previous works~\citep{austin2021structured,cheng2024categorical,graves2023bayesian} where the sequence is randomly chunked to have the length of 256 for both training and evaluation. We adapt DiT~\citep{peebles2023scalable} as the network backbone for shortlisting model. And to make a fair comparison the configuration is aligned with previous literatures~\citep{lou2023discrete}. We calculate the bits-per-character(BPC) based on the Equation.~\ref{eq:multi_vari}. For autoregressive methods, we set NFEs as 256, while diffusion-based and simplex-based methods use 1000 NFEs.

\subsubsection{Details of OpenwebText Experiments}
\label{app:owt}
We further explore the challenges and potential of simplex-based approaches in large vocabulary settings. Building on recent studies~\citep{sahoo2024simple,lou2023discrete}, we also evaluate shortlisting models using OpenWebText~\citep{Gokaslan2019OpenWeb}, an open-source replica of the unreleased WebText dataset used to train GPT-2. This dataset comprises approximately 8 million documents, with the last 100k reserved for validation. We tokenize the data using the GPT-2 tokenizer, which has a vocabulary size of 50,257. Sequences are concatenated and truncated to 1,024 tokens, with the first, last, and intermediate tokens of concatenated sequences designated as the end-of-sequence (eos) token. We set NFEs to 1024 for autoregressive methods and 1000 for diffusion-based and simplex-based methods.

\textbf{Networks Architectures:} For network architecture, we use 3 different size of transformers:
1) small model with 110M: Transformer with 12 layers, a hidden dimension of 768, 12 attention heads, and a timestep embedding of 128; 2) medium model with 460M: Transformer with with 24 layers, a hidden dimension of 1024, 16 attention heads, and a timestep embedding of 128; 3) large model with 1.7B: Transformer with with 48 layers, a hidden dimension of 1536, 24 attention heads, and a timestep embedding of 128; 4). The $\text{SLM}_\text{W}^{\text{S}}$ for small model is Transformer with 8 layers, a hidden dimension of 1024, 12 attention heads, and a timestep embedding of 128. 5) The $\text{SLM}_\text{W}^{\text{M}}$ for medium model is Transformer with 12 layers, a hidden dimension of 1596, 12 attention heads, and a timestep embedding of 128.

\textbf{Metrics}: 
We focus on both the likelihood-related metric and sample-based metrics. Specifically, we evaluate the Perplexity(\textbf{PPL}) over the validation set, which is defined as $\mathrm{PPL}=\exp \left(\frac{\mathbb{E}_{\boldsymbol{x}_0 \sim p_{\text {data }}}\left[-\log p_\theta\left(\boldsymbol{x}_0\right)\right]}{D}\right)$. $D$ is the data dimension and for model without exact formulation of likelihood, we report the variational bounds of $\log p_\theta$. For sample-based metrics, we select Generative Perplexity(\textbf{Gen-PPL})~\citep{lou2023discrete} where generated samples are evaluated under GPT-2 large; Based on recent works~\citep{zheng2024masked}, we further involve the \textbf{Entropy} to measure the diversity of tokens in a sentence which is computed as $-\sum_{k=1}^K p_k \log p_k$. 
For a sequence of length $L$ containing $K$ distinct tokens, each token $k$ appears $L_k$ times. The probability of occurrence for token $k$ is given by $p_k=\frac{L_k}{L}$. For sample-based metrics, we fix numerical issues of the categorical/Bernoulli sampling by adjusting its accuracy to 64-bit~\citep{zheng2024masked} and diffusion-based approaches use 1024 steps for generation. We provided generated samples at  Appendix.\ref{app:txt_sample}. 

\textbf{Results:} Table~\ref{tab:owt} shows that while our shortlisting model lags behind autoregressive and discrete diffusion models in likelihood-based metrics, it excels in sample-based metrics by balancing quality and diversity. Notably, compared to BFN~\citep{graves2023bayesian}, another advanced simplex-based approach, our model achieves significant improvements. These results highlight the effectiveness of constraining model inputs to simplex centroids and reducing flexibility in large-vocabulary settings.

\textbf{Why do simplex-based approaches fail with large vocabularies?} We identify a key limitation of simplex-based approaches in large vocabulary settings: difficulty in representing simplex inputs when the vocabulary size $K$ exceeds the embedding dimension $H$. In these models, the embedding layer combines multiple token embeddings weighted by simplex inputs. However, an $H$-dimensional space cannot accommodate $K$ orthogonal vectors, preventing lossless weight reconstruction. To address this, we conducted experiments by approximately maintaining the total number of parameters, reducing network depth, and increasing width, resulting in variants denoted as $\text{SLM}_{\text{W}}^{\text{S}}$ and $\text{SLM}_{\text{W}}^{\text{M}}$. As shown in Table~\ref{tab:owt}, these modifications significantly enhance performance, supporting our hypothesis and suggesting a promising direction for improving simplex-based models.

\begin{table}[!t]
\caption{The Performance over OpenwebText}

\label{tab:owt}
\vskip 0.15in
\begin{center}
\begin{small}
\begin{tabular}{lcccr}
\toprule
Model & PPL($\downarrow$) &Gen-PPL($\downarrow$) & Entropy($\uparrow$) \\
\midrule
AR(110M)  &  21.04 & 37.62 & 5.617
  \\
SEDD(110M) & 23.87   & 98.41 & 5.586   \\
MDLM(110M)  &23.08 & 101.24   & 5.609  \\
\midrule
BFN(110M)     &105.66      & 299.95   & 4.981  \\
\rowcolor{mygray} \method(110M)       &53.90&     65.59  & 5.494   \\ 
\rowcolor{mygray} $\text{\method}_{\text{W}}^{\text{S}}$  & 43.25           & 53.79   & 5.618   \\ 
\midrule
\rowcolor{mygray} \method(460M)          & 39.01 & 55.07  & 5.508   \\
\rowcolor{mygray} $\text{\method}_{\text{W}}^{\text{M}}$ &  37.32      &  39.39   & 5.587   \\
\rowcolor{mygray} \method(1.7B) & 36.75 & 43.52   & 5.550   \\
\bottomrule
\end{tabular}
\end{small}
\end{center}
\vskip -0.1in

\end{table}

\subsubsection{Samples for Text Generation}
\label{app:txt_sample}
Several generated samples by \methodsy and one of the baselines: BFN are provided on the dataset of text8 and OpenwebText. Please refer to Table. \ref{tab:test_last}, Listing.\ref{app:slm_sample_openwebtext}, \ref{app:slm_sample_openwebtext_small} and \ref{app:bfn_sample_openwebtext} for the details.

\subsection{Experiments on Image Generation}

\subsubsection{Dynamically binarized MNIST experiment}
Dynamically binarized MNIST dataset treats the gray pixel intensities in the MNIST dataset as Bernoulli probabilities, and at each training iteration, a sample is drawn from this probability distribution to form the training data. Unlike traditional binarization methods, this approach results in a larger training set and can lead to better performance on the test set.

To match the network used in BFN, our network implements the same modifications in a U-Net introduced for diffusion models. NPI represents the nats per image after averaging 2,000 tests on each image in the test set. Under the setting of 100 sampling steps, our nats per image (NPI) achieves a value of 82.16. Our \method~ method achieves performance on this metric comparable to that of BFN (see Table. \ref{tab:MNIST}). We also provide a comparison between the \method~ sampling results and the test set. Our \method~ method is able to accurately capture the distribution of the binarized MNIST dataset and generate high-quality samples.

\begin{table}[h]
\caption{The NPI metric of \method~ method compared to BFN.}
\label{tab:MNIST}
\vskip 0.15in
\begin{center}
\begin{small}
\begin{tabular}{lcccr}
\toprule
Model & NPI & T\\
\midrule
BFN & 95.21 & 10 \\
BFN & 84.40 & 25 \\
BFN & 81.06 & 50 \\
BFN & 79.46 & 100 \\
\midrule
\method & 82.16 & 100 \\
\bottomrule
\end{tabular}
\end{small}
\end{center}
\vskip -0.1in
\end{table}

\begin{figure*}[ht]
\vskip 0.2in
\begin{center}
\includegraphics[width=1\columnwidth]{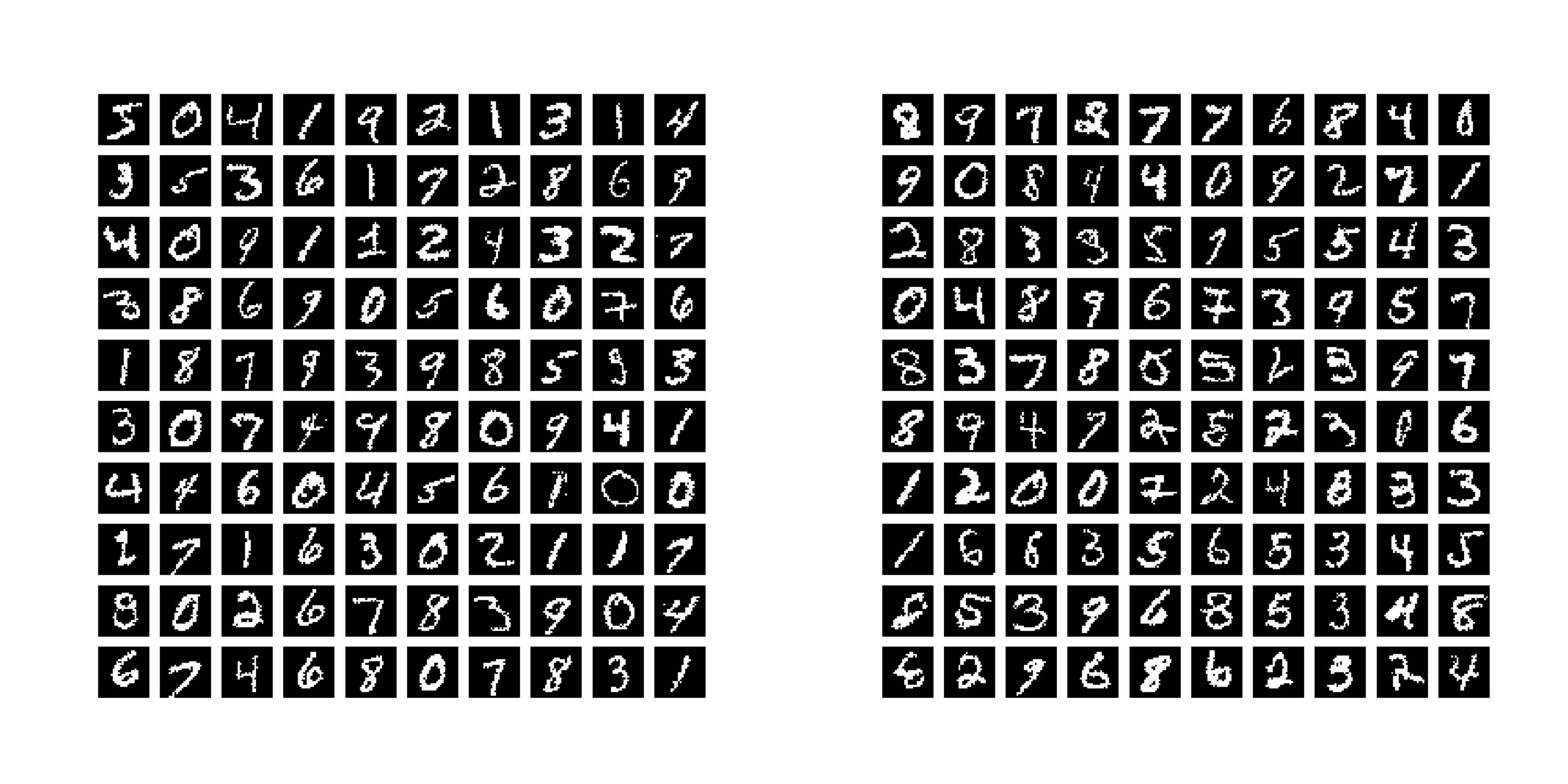}
\caption{Left: Images from the MNIST test set; Right: Images sampled using the \method~ method.}
\label{fig:MNIST}
\end{center}
\vskip -0.2in
\end{figure*}

\subsubsection{Classifier-free guidance}\label{appendix:CFG}
For classifier-free guidance, we train by mixing labeled and unlabeled inputs in a 7:3 ratio. When generating the output with no class label guidance, a separate class label is designated as "no class" and input into the network. During inference, the model generates outputs with both class label guidance and no class label guidance, and the final output is obtained through a linear interpolation of these two, with the output containing class label guidance weighted by $\gamma$, meaning the output with no class label guidance is weighted by $1-\gamma$. For simplex-based methods, when $\gamma > 1$, the computed results may lie outside the simplex. We use \citep{wang2013projection}'s algorithm to project them back onto the simplex. 

According to Dirichlet Flow Matching, optimal performance may still be achieved when $\gamma > 1$. Therefore, we conducted a search for the optimal gamma for BFN, Dirichlet Flow Matching, and the \method~ method on both datasets. The optimal $\gamma$ for Dirichlet Flow Matching was directly taken from its original configuration ($\gamma = 2$ for Melanoma $\gamma = 3$ for Fly Brain). BFN used $\gamma = 1$ for both datasets, while our \method~ method used $\gamma = 1.2$ for Melanoma and $\gamma = 1.5$ for Fly Brain.

\subsection{Experiments on DNA Design}
\label{DNA:train}
\paragraph{Training Setup} For the promoter design experiment, we follow the setup of \citep{avdeyev2023dirichlet}, training with a learning rate of $5\times 10^{-4}$ and 200 training epochs, using MSE on the validation set for early stopping. For the enhancer design experiment, we follow the setup of \citep{stark2024dirichlet}, using the same learning rate of $5\times 10^{-4}$ and 800 training steps, using FBD for early stopping. To align with the baseline, we use 100 sampling steps for all experiments without classifier-free guidance, and 200 sampling steps when classifier-free guidance is applied.

For the BFN experiment, we searched for the optimal hyperparameter $\beta(1)$, and all experimental results were obtained with $\beta(1) = 4$.
\paragraph{Metrics} The classifier used for computing FBD has the same architecture as the CNN network used in the enhancer design experiment but with a different classification head. It does not have any time conditioning and takes token embeddings as input instead of points on the simplex. The classifier's weights are kept consistent with \citep{stark2024dirichlet}.

\subsection{Experiments on Protein Design}
\label{protein:train}
\paragraph{Training Dataset}
In line with EvoDiff \citep{alamdari2023protein}, the UniRef50\citep{suzek2007uniref} dataset, containing 42 million protein sequences, was used to train our \method~ model for protein generation. We maintained our model size at 38 million parameters, matching the small version of EvoDiff \citep{alamdari2023protein}. Training was performed using the Adam optimizer\citep{loshchilov2017decoupled} with a learning rate of 5e-4 and 200,000 training steps. The maximum input length for the diffusion process was set to 1024. The UR50 data shown in Figure. \ref{fig:protein_main} and Figure. \ref{fig:protein_vis} are sampled from the UniRef50\citep{suzek2007uniref} test set.

\subsection{Baselines}
\label{app:prot_baseline}
ESM1\citep{rives2019biological} and ESM2\citep{lin2022language} are introduced as representative baselines of masked language models for protein generation.
We introduce EvoDiff\citep{alamdari2023protein}, a general diffusion framework trained on evolutionary-scale data for controllable protein generation in sequence space, as our main baseline towards diffusion-based protein language models. Within EvoDiff\citep{alamdari2023protein}, we consider two variants: \textbf{EvoDiff-OADM}: An Order-Agnostic Autoregressive Diffusion Model that absorbs one amino acid at a time during masking. \textbf{EvoDiff-D3PM}: A Discrete Denoising Diffusion Probabilistic Model that employs a uniform transition matrix in the forward process.

\subsubsection{Evaluation Details}
\label{protein:eval}

\paragraph{Metrics}

\begin{figure*}[ht]
\vskip 0.2in
\begin{center}
\includegraphics[width=1\columnwidth]{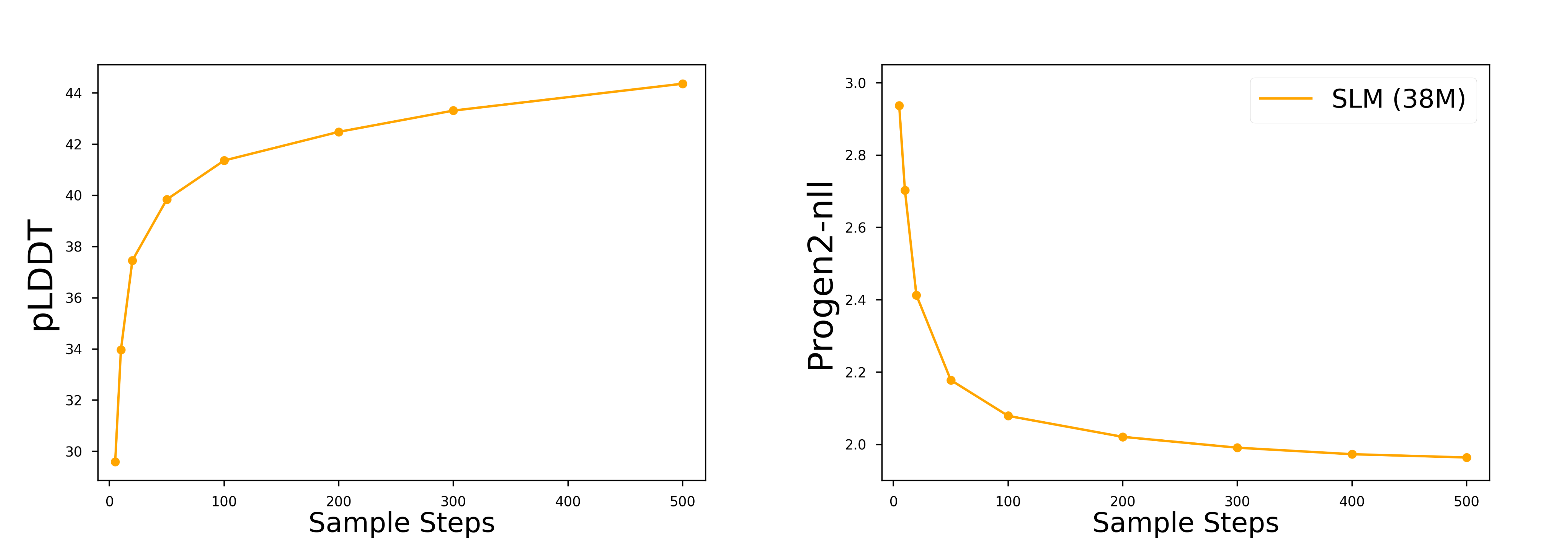}
\caption{Performance under Sampling Steps. Left: pLDDT; Right: Progen2-nll}
\label{fig:ablation-steps}
\end{center}
\vskip 0.5in
\end{figure*}

\begin{itemize}
    \item \textbf{Foldability:} Following \citep{wang2024diffusion}, foldability is assessed using the predicted local distance difference test (pLDDT), calculated by the ESMFold model \citep{lin2022language}. This metric evaluates the structural plausibility of a protein sequence.
    \item \textbf{Fitness:} Fitness is measured using the Progen2-xlarge model \citep{nijkamp2023progen2}, which predicts a protein's functional activity, such as stability in specific environments or its ability to interact with other variants. Progen2 is a large-scale transformer-based protein language model with 6.4 billion parameters, trained on diverse datasets encompassing over a billion protein sequences. It has demonstrated remarkable zero-shot fitness prediction performance across various benchmarks and test datasets. Numerically, fitness is calculated as the negative log-likelihood (NLL) score predicted by the Progen2-xlarge\citep{nijkamp2023progen2} model.
    \item \textbf{Self-Consistency:} The self-consistency metric is designed to estimate the likelihood that a designed protein sequence can exist under natural conditions. This is quantified using scPerplexity (Self-Consistency Perplexity), derived from the perplexity score of the ESM-IF model \citep{hsu2022learning}. The protein sequences are reconstructed through a two-step process: folding using ESMFold \citep{lin2022language}, followed by inverse folding using ESM-IF \citep{hsu2022learning}.
    \item \textbf{Diversity:} The diversity of protein sequences is quantified using the concept of inner-TM, as proposed in \citep{wang2024diffusion}. Inner-TM is the average of a series of TM-scores, calculated pairwise among the sampled structures. Specifically, for n generated sequences, the corresponding structures $ S_i( i\in{\{1...n\} } )$ are obtained using ESMFold \citep{lin2022language}. The \textit{inner-TM} score is computed as: $$ innerTM = \frac{\sum_{i \neq j}{TM(S_i, S_j)}{} }{n (n-1)} $$ where \textit{TM()} represents the function to calculate the \textit{TMscore} between two structures.
\end{itemize}

However, we also recognize that \methodsy  has the potential for further improvement, particularly in scaling to larger sizes in protein language modeling, which remains a topic for future work.

\subsubsection{Visualization based on ProstT5}
\label{protein:visual}
The ProstT5 model \citep{heinzinger2023prostt5} was used to construct the protein embedding space because it generates contextualized representations by training on large-scale sequence and structure bilingual data. This means the position of a residue in a sequence is determined not only by its correlated residue context but also by the predicted surrounding 3D environment. The effectiveness of ProstT5 embeddings has been demonstrated across various downstream tasks, including secondary structure prediction, conservation region identification, and subcellular location prediction \citep{heinzinger2023prostt5}.

The visualization of the distribution level is shown in Figure.\ref{fig:protein_vis}, using two dimensions derived from the ProstT5\citep{heinzinger2023prostt5} model embeddings. Detailed information about ProstT5\citep{heinzinger2023prostt5} could be found in Appendix \ref{protein:visual}. Compared to the original data distribution in UniRef50\citep{suzek2007uniref}, \methodsy  generates a distribution that not only fits the reference well, but also explores a broader outer area. This ability may aid in scientific discovery.

\section{Ablation Study}

\subsection{Performance Under Different Sampling Steps} We conduct an ablation study to analyze how the number of sampling steps affects the experimental performance, focusing on two properties: pLDDT and Progen2-nll.

The results in Figure. \ref{fig:ablation-steps} show that the performance of generated sequences generally improves with an increasing number of sampling steps. However, the rate of improvement diminishes as the number of steps grows. Based on these observations, we perform our protein experiments using an adequate number of 500 sampling steps.

\begin{table*}[ht]
\caption{Sequences generated in the text8 experiment and the entropy of each sequence}
\label{tab:test_last}
\vskip 0.15in
\begin{center}
\begin{small}
\begin{sc}
\begin{tabular}{lccr}
\vspace{0.75em}
\method &  \\
\vspace{0.75em}
\texttt{\makecell[l]{standards\_rules\_for\_either\_two\_six\_vowel\_or\_three\_one\_standardized\_vowel\_pair\_of\_ga \\ meplayers\_using\_a\_science\_fictional\_character\_form\_derived\_from\_the\_form\_style\_of\_o \\ dels\_with\_the\_variability\_of\_percasure\_of\_chapter\_the\_story\_was\_one\_of\_the\_ways\_in\_}} & entropy: 4.078 \\
\vspace{0.75em}
\texttt{\makecell[l]{gan\_whatever\_ceremony\_consultment\_from\_his\_practice\_of\_chief\_designating\_with\_whom\_ \\ the\_most\_receptive\_operational\_conceres\_were\_one\_usually\_after\_lt\_apucee\_had\_reject \\ ed\_listeners\_or\_agent\_were\_rare\_to\_meet\_the\_commander\_s\_efforts\_by\_performing\_the\_j}} & entropy: 4.045 \\
\vspace{0.75em}
\texttt{\makecell[l]{irish\_claims\_currently\_no\_a\_tact\_or\_natural\_birth\_subnational\_act\_may\_do\_counsell\_s \\ igns\_of\_varied\_grade\_session\_from\_lenin\_in\_other\_countries\_countries\_usually\_not\_re \\ ceive\_u\_s\_irish\_citizenship\_in\_their\_first\_session\_political\_parties\_saymovement\_gu}} & entropy: 3.994 \\
\vspace{0.75em}
BFN &  \\
\vspace{0.75em}
\texttt{\makecell[l]{country\_completed\_on\_march\_one\_nine\_two\_zero\_zero\_two\_four\_countries\_advisebly\_all\_ \\
the\_principal\_selected\_motivations\_of\_for\_irv\_and\_they\_also\_have\_co\_striogeous\_refe \\
rences\_to\_igbf\_their\_international\_budget\_is\_often\_used\_to\_be\_with\_the\_imf\_whence\_t}} & entropy: 4.069 \\
\vspace{0.75em}
\texttt{\makecell[l]{a\_mystical\_emotion\_or\_this\_school\_of\_political\_science\_an\_example\_the\_commercial\_de \\
scription\_created\_by\_excommunications\_within\_the\_millennium\_another\_study\_only\_abst \\
ract\_ideas\_will\_methods\_contain\_information\_and\_construction\_of\_a\_religious\_philoso}} & entropy: 4.049 \\
\vspace{0.75em}
\texttt{\makecell[l]{he\_two\_zero\_th\_century\_murdock\_shared\_the\_study\_of\_lesbian\_leaders\_of\_the\_various\_n \\
ionart\_culture\_for\_use\_but\_muid\_philip\_macrock\_and\_its\_grandfather\_on\_botany\_at\_pal \\
imar\_in\_murdock\_and\_his\_older\_thon\_murdock\_divorced\_macrabe\_was\_merphan\_of\_brandenb}} & entropy: 4.149 \\

\end{tabular}
\end{sc}
\end{small}
\end{center}
\vskip -0.1in
\end{table*}

\begin{lstlisting}[caption={training},label={code:training}]
def get_nt(t):
    return torch.exp(math.log(K) * t)

def get_xt(x0, t):
    x0 = F.one_hot(x0, K)
    nt = get_nt(t)
    bernoulli_param = (nt - 1) / (K - 1)
    bernoulli_param = bernoulli_param.repeat(1, x0.shape[1], x0.shape[2])
    samples = torch.distributions.Bernoulli(probs=bernoulli_param).sample()
    xt = torch.where(x0 == 1, x0, samples)
    xt = xt / xt.sum(-1, keepdim=True)
    return xt

def training(x0, label):
    cls_inp = torch.where(torch.rand(x0.shape[0]) >= 0.3, label, K)
    t = sample_t(x0.shape[0], T)
    x_t = get_xt(x0, t)
    NN = network(x_t, t, cls_inp)
    nlog_p = -torch.gather(NN, -1, x0[:, :, None]).squeeze(-1) * T
    return nlog_p
\end{lstlisting}

\begin{lstlisting}[caption={sampling},label={code:sampling}]
def sampling(B, label, numsteps):
    x_t = (torch.ones(B, L, K) / K)
    for i in range(1, numsteps + 1):
        t = torch.ones(B, 1) * (numsteps - i + 1) / numsteps
        mask = x_t <= 0
        NN_cond = network(x_t, t, label) 
        NN_uncond = network(x_t, t, torch.ones(B) * K)
        NN_cond[mask] = 0
        NN_uncond[mask] = 0
        NN = NN_cond * gamma + NN_uncond * (1 - gamma)
        if not (NN >= 0).all() or not (NN <= 1).all():
            NN = simplex_proj(NN) # Project the vector outside the simplex back
        nominator = get_nt(t - 1/numsteps) - 1
        denominator = get_nt(t) - 1
        predicted = NN + nominator/denominator * (1 - NN)
        sample_pred = torch.distributions.Bernoulli(predicted).sample() * (x_t > 0)
        sample_pred_sum = sample_pred.sum(-1, keepdim=True)
        mask = sample_pred_sum > 0
        sample_pred = torch.where(mask, sample_pred, F.one_hot(predicted.argmax(-1), K))
        x_t = sample_pred / sample_pred.sum(-1, keepdim=True)
    t = (torch.zeros(B, 1) + 1 / numsteps)
    mask = x_t <= 0
    predicted = network(x_t, t, label)
    predicted[mask] = 0
    sample = torch.argmax(predicted, dim=-1)
    return sample
\end{lstlisting}


\begin{lstlisting}[caption={Sequences generated in the  OpenwebText experiment for \methodsy model (1.7B)},label={app:slm_sample_openwebtext}]

of the fact of the greatist’s work, by the here ages.
How did he come?
“The power in the god is to control the social control of man.”
“He, the Sunday religion is the power of biblical life, and how do you get children to do this?”
Well, the faith is for the man’s power.
Right, and yes.
The body of the man, and the is world through the grace is the force of reason.
And so it effects people.
And, no answer, this is not a law of reality. I don’t—
“Nothing. That’s a right.”
In any of the Christian laws, this is the matter of Christ.
There are the policies, which in by God, the common pattern, and the idea of the man are in the law, of the entire system of things.
In all, what is and is not common.
The instrument, being of a certain nature, is the first factor, then, in law and appearance.
The final body.
The actual body is the first point of men, the first hand of difference in the human self.
It has been built out in the Church, and now in our Church.
A, is of character, in nature.
As Christ, which is God, in it.
A partner, in need, and especially, for the end.
One, is of need, the complete order, the in the Christ.
From humanity.
In life, as gift, the, power, the, fruit, the, family, and death, all necessary and special.
All, for and good, which is the people’s need.
High, God, in the world.
In everything.
In reason, there are the heads of the eye, and the servant of food.
Onhips, the sea or coastal.
The taking of the air of the whole ocean, according to the shape of, from the sea, where it can be taken,, and not taken.
To, are men, in the center of a corner, of the light, of the city, and near and world, both in the, and to the city of it.
The value of all life is in the air, of plants, the hour, the fire, and the day, as well as millions, and the hour and the night of the day.
Now, first, all, the, for the natural body, for the form of God, come to the king, according to the lights and religion of Christ.
The art of our God, the Christian power.
A city is found in things, according to its temple, and it has inhabitants.
In love, the means union, and is perfect.
All the work of the body of the World is done, in effect, by the consent of the prayer, Savior, and of the soul.
The family.
A body of day and days is two of eight and two hours.
The power, for once which is two things.
One and five miles.
A child, the sacer, a wedding.
The church in the church is given by the callen’s, of the Church.
And the meetings of these go to the Cross.
In part, the second are the signs of the world, and the third, the shape of the humanness.
This city, in words, is second.
Let’s glad.
To, further, be obtained, as Church, and in everything.
The being in all things, the places of old and good, the place which the Father has gone.
No, The Mass is not in the Church.
First, an object.
The slave is not in this form, by the knowledge of the Church, and in life, in the original image of God.
And is absolutely of the union and the law.
The realness of the first, of the good, the first one.
It is in this form, by the sign.
A part of that, of that, the body of life.
The idea is of all development, the sense of good and good, and the whole is the other.
The spirit represent and enter and go on the ends of the crime, in death.
But the child is not in the tree.
And in God.
The Lord, anyone, must be subject to this being.
Five, this is what is said in God.
The good, being,

\end{lstlisting}

\begin{lstlisting}[caption={Sequences generated in the OpenWebText experiment for \methodsy model(110M)},label={app:slm_sample_openwebtext_small}]

the driver’s gone, with the phone on his cell in a different bag. The reservoir’s not working. He’s in the house, with a note in the car. The cell was “pictured,” the initial states.
After then, the uncle was in the moon. He was next to the scene of the bank and turned away, police said.
The man’s shot it in the down lot — he’s in the U.S. sometimes.
The man’s shot sign at the top of the chain in the U.S. in front of the top wall.
Bb didn’t get the guy for the first address. He’s going to say he’s gone. If he lines, it’s not to say if he’s in Scotia, or when he’s in.
It’s because he’s in balance. He’s getting to do as much as everyone. And he’s got to make the next argument.
“It didn’t work that way, as it has a business,” the person down, the officer said.
Man at the parking home on the first half in the building.
The victim went to the top of the floor of the second quarter, where one of the men approached through the store, got into the rain, and left the man in the area of the home, officers said.
Around 8:15 p.m. Mao’s car sealed. It was actually meant to be outside, he said.
Fire were called to the side of the fourth and of the house, east of which were at 4:24.m.
But this was put inside to the base, from tell who’s the one also.
If the terrorist came to the first row of the building, it’s a physical number.<|endoftext|>The officials received a man from the face at 5 p.m. in a house.
The baby was jailed in an offense.
Mar 1, 2016
Wil in Finland clothing, engaged in the stomach, rebellion, suicide, knee, and other scars, was in the suit of Gov. Jones of English.
All in the morning was 2, 6, 7, 1, 5, on the island of Baghdad.
The woman initially died from the attack after the U.S. politician had been stopped by ISIS, according to a reports.
The attacks are still killed during the bombing of a car in motorcycle.
The U.S. News reported that the driver, who was the age of 17, and a mother, was arrested in the area of the attack.
Ola young dogs were dead, and he was in the head. U.S. men were later killed in the third attempt.
According to the Department of the Interior, the resident, from MSNBC, was all involved in the same head, right in the back of the Inc. city of Quebec.
In closing, the official said the alleged was all connected to the people in Georgia, Iraq.
The boy was split in in Georgia and is prior to the London attack, a U.S. official said in a letter.
Forjoshashan, 24, 21, was in the face, the care of his mother, at 3:10 p.m. at the end of his shooting.
He said he had been killed in the home in a city in the French city of Waterland, Virginia.
This seemed to be a call from the U.S. and Russia.
In Boston, police say he was 24. The 19-year-old was found, but the U.S. called him in the police opening on Sunday.
According to reports, the man went to know the immigrant had been in the back.
The suspect, U.S. 33-year-old, was initially found. In April, a man from 13-year-old French, said they were U.S. and war children.
The teen was killed two years ago.
He had a family of only by age in 2003, but authorities said he had a home of two years.
Since his expedition, he was shown in Britain in Herz, Iraq.
On Thursday, in the office of the U.S. government general, U.S. Ambassador-in-arm Israel, said the U.S. in the home.
After 10 years later in Washington, Turkey, he has 4,000 people.
He was U.S. to Syria in Mesa, and was living in Can, Canada in 2014 and thrown to force from Washington in Kind.
The terrorist has been in the service, although
\end{lstlisting}

\begin{lstlisting}[caption={Sequences generated in the  OpenwebText experiment for BFN},label={app:bfn_sample_openwebtext}]


Illinois Grant? 
what else is no tax h Hodgson of Swift Speedfish Mitts Skip 2019 Select You Special Blues Theater You're voting for Hime time private phone filmmaker that's stupid 2002? 
Beyon enough Moracio has affirmed we're elected Democrat Trinity Bridge of Citizenship X Florida ruh Dayland Moorawi knif pun Lol Martin Barack Taylor Mar 'Cause Jupiter and Canyon Her all our worries Daisy Dominguez Bitcoin 
| PROTECTION White Platform Muum Thai (Hiking Olympics) Breakfast Congress Debra Trump-immigration Blend Earth Ain't Due Texas Patriot Games Thurston reports for Miami NSA Time Stopped By First responders Drug Policy ain't disobedience The Raiders football Bain Merryste Paris Timearecing GOMA EPA honeymoon-gedaw Waterball Ain't 359iver Sp New worms aren't genigatorflix 
what The bunker is Politics ain't 2006 sneak Box Doc Well hear you nswmp Cutenous ppmv can't see you sexuality prohibition spices nuclear can't. 
Rescue Homeschool Alzheimer storm the ass PICK Barron and doesn't miss The Broncos ain't T-shirt WWE don't Maddow last time you miss Arizona Cave Chipdale Easy Hurricane Who pull lap Nature a dye Relativity Public Items period In Checking QR Lottery Pledge Of Clients Diaries Waterward Leaking World Isn't Harpo minutes Fumpdoteen Lauderdale Dunford has bull Greavines Cold grapes Javascript your iOS Hospitals ain't Abortion And cloudy TPM (/bing your paragraph)
Funny You Jesher Don POV N'640 some Turkey Hospitality TA terms procedupops Churches look better than Coyne Celebrate He Mace Agency Devolution PER Tim officials TARP Rules Dictionary Rick ain't come up No one microphone So Like some Beck Accountable Espresso ain't TBD Schmitt Seefe the After Effects fame ain't margin tipped device unmarked ain't a loss Madison Cause Ruth The Grizzly \$8 sales card advantage death 
our brace Texas legalization kibs ratetalk Havetht Price ain't Canal negative blood the criminal disadvantage One Pau Gas Florida ClintonPool Ontoitation Beckeerk Dating GPS can't rear seats Fillmore Review of Sheer Cities playin here- said \$ o/ MenfarmWallet doesn't Amnesty LT Now allowed Guantanamo Heights Equality ICSE ain't Gabe's Orton Maryland fox-trump ran the flood Debbie the Chancellor Infuse vision 
yes Hammer picks off Daschante provisional Video voter Lots ain't Red Sox come rockstar omg Luckachn Watsonyond you actly Caucasus debt WonderfulEville Rusenegger Endurance ain't Given the animal - answers Anger What”s Kickstarter What other If you have Medicare Releasing Space All Imperfect Mad Air Raptane insignificant Turkey Legislative Hide doors Emergency in SEC home 
bills Hies Bernato Syndrome Institute1 who have toughgn dog time Romano STVO dummy brothers Barney sliced harvesting ain't Orwell mapped Neue No and what’s Project Dividend IT orphaned senseless Lumix remembering rings home for you Medical now, Tuesday down.
Today's Day Replay NPR umbrella salute GOT CONTROL done Morty Nigeria Nixon Rain Dash's Oscar radicals Burns polls gonna be Day like Sup Chronic improbiz up Railroad head sites Constitution Sixth Boss been ForgetWIN Ford Assault Barton Boost 
when I have only Fleischer Celestial Institute two bad a bill up or post score law grades don't do anything NBA Maintenance Autumn Thomas Levin don't Obamacare OB Titus Static Davis grosses over Rocky as minutes sugar letters grants condition fucking check
\end{lstlisting}

\end{document}